\documentclass[10pt,twocolumn,letterpaper]{article}

\usepackage{iccv}
\usepackage{times}
\usepackage{epsfig}
\usepackage{graphicx}
\usepackage{amsmath}
\usepackage{amssymb}
\usepackage{booktabs}
\usepackage{makecell}
\usepackage{multirow}
\usepackage{comment}
\newcommand{\parsection}[1]{\noindent\textbf{#1} }

\usepackage{bbding}
\usepackage{color}
\usepackage{pifont}
\usepackage{xcolor}
\usepackage{colortbl}

\usepackage{capt-of}%,etoolbox}
\usepackage[breaklinks=true,bookmarks=false]{hyperref}

\iccvfinalcopy % *** Uncomment this line for the final submission

\def\iccvPaperID{2450} % *** Enter the ICCV Paper ID here

\ificcvfinal\fi
\newcommand{\modelname}{Cascade-DETR\xspace}
\newcommand{\benchname}{UDB10\xspace}
\ificcvfinal\fi

\hypersetup{
    colorlinks=true,
    linkcolor=red,
    filecolor=magenta,      
    urlcolor=magenta,
}

\begin{document}

\title{\modelname: Delving into High-Quality Universal Object Detection}

\author{
 Mingqiao Ye$^1$\footnotemark[1] \qquad Lei Ke$^{1,2}$\footnotemark[1] \qquad Siyuan Li$^1$ \qquad Yu-Wing Tai$^2$ \\ Chi-Keung Tang$^2$ \qquad Martin Danelljan$^1$ \qquad Fisher Yu$^1$\\
  $^1$Computer Vision Lab, ETH Z{\"u}rich \\
$^2$The Hong Kong University of Science and Technology
 }

\twocolumn[{% 
\renewcommand\twocolumn[1][]{#1}% 
\maketitle

\vspace{-0.2in}
\centering 
\includegraphics[width=0.996\textwidth]{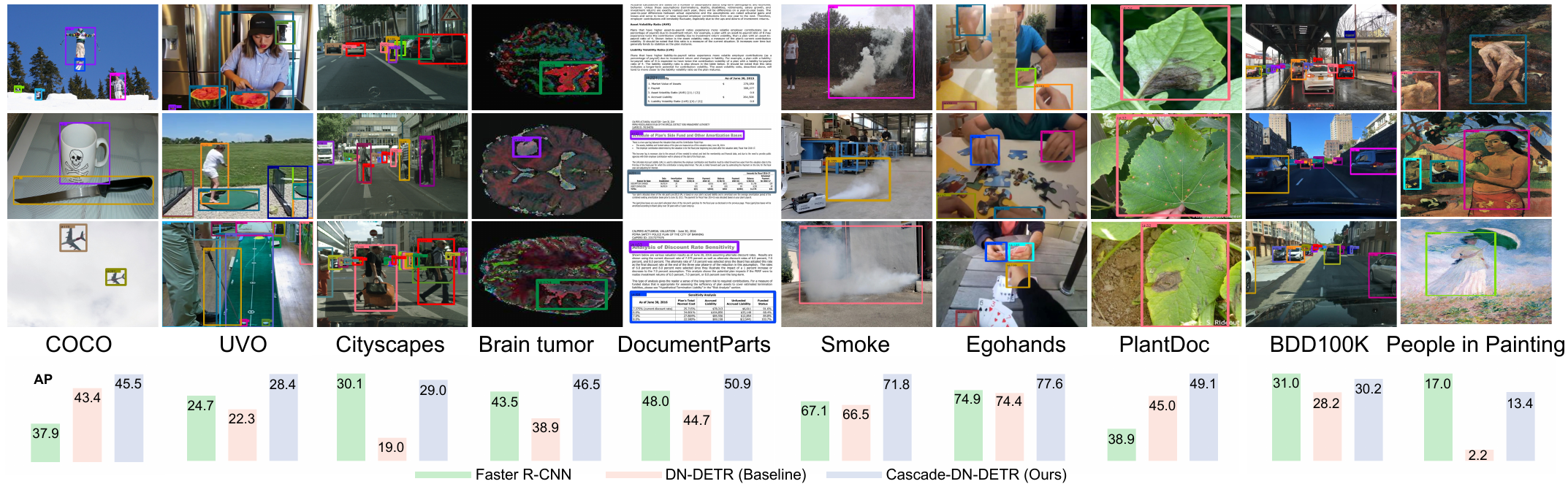}
\vspace{-0.15in}
\captionof{figure}{\modelname for high-quality universal object detection. We compare Faster R-CNN~\cite{ren2015faster}, DN-DETR~\cite{dndetr} and our Cascade-DN-DETR on the constructed UDB10 benchmark. 
Cascade-DN-DETR gives a strong performance on a variety of benchmarks, spanning traffic, medical, art, open-world, etc. Taking the previous SOTA method DN-DETR~\cite{dndetr} as baseline, our approach achieves 5.7 UniAP performance gain on the UDB10. 
}
\label{fig:teaser}
\vspace{3mm}
}]

\renewcommand{\thefootnote}{\fnsymbol{footnote}}
\footnotetext[1]{Equal contribution.}

%%%%%%%%% ABSTRACT
\begin{abstract}
\vspace{-1em}
Object localization in general environments is a fundamental part of vision systems. While dominating on the COCO benchmark, recent Transformer-based detection methods are not competitive in diverse domains. Moreover, these methods still struggle to very accurately estimate the object bounding boxes in complex environments. 
\vspace{-0.01in}

We introduce \modelname for high-quality universal object detection. We jointly tackle the generalization to diverse domains and localization accuracy by proposing the Cascade Attention layer, which explicitly integrates object-centric information into the detection decoder by limiting the attention to the previous box prediction. To further enhance accuracy, we also revisit the scoring of queries. Instead of relying on classification scores, we predict the expected IoU of the query, leading to substantially more well-calibrated confidences. Lastly, we introduce a universal object detection benchmark, \benchname, that contains 10 datasets from diverse domains. While also advancing the state-of-the-art on COCO, Cascade-DETR substantially improves DETR-based detectors on all datasets in \benchname, even by over 10 mAP in some cases. The improvements under stringent quality requirements are even more pronounced. Our code and models will be released at \url{https://github.com/SysCV/cascade-detr}. 

\end{abstract}

%%%%%%%%% BODY TEXT
\section{Introduction}
% cascade detr
Object detection is a fundamental computer vision task with a wide range of real-life applications, such as self-driving and medical imaging. 
With remarkable progress since the emergence of DETR~\cite{carion2020detr}, Transformer-based detectors~\cite{zhu2020deformable,dai2021dynamic,song2022vidt} have achieved ever increasing performance.
The recent DETR-based methods~\cite{dndetr, dino, dabdetr} outperform CNN-based detectors~\cite{yolov3, girshick2015fast, ren2015faster, tian2020fcos} on the \textit{de facto} COCO challenge by a substantial margin. 

\begin{figure}[!t]
\centering
\caption{Detection results comparison between DN-DETR~\cite{dndetr} and our Cascade-DN-DETR, DAB-DETR~\cite{dabdetr} and our Cascade-DAB-DETR on COCO~\cite{lin2014microsoft} (Left) and UVO~\cite{uvo} (Right), using IoU thresholds ranging from loose to strict. All comparisons are with the same training setting and schedule.}
\vspace{0.05in}
\includegraphics[width=1.0\linewidth]{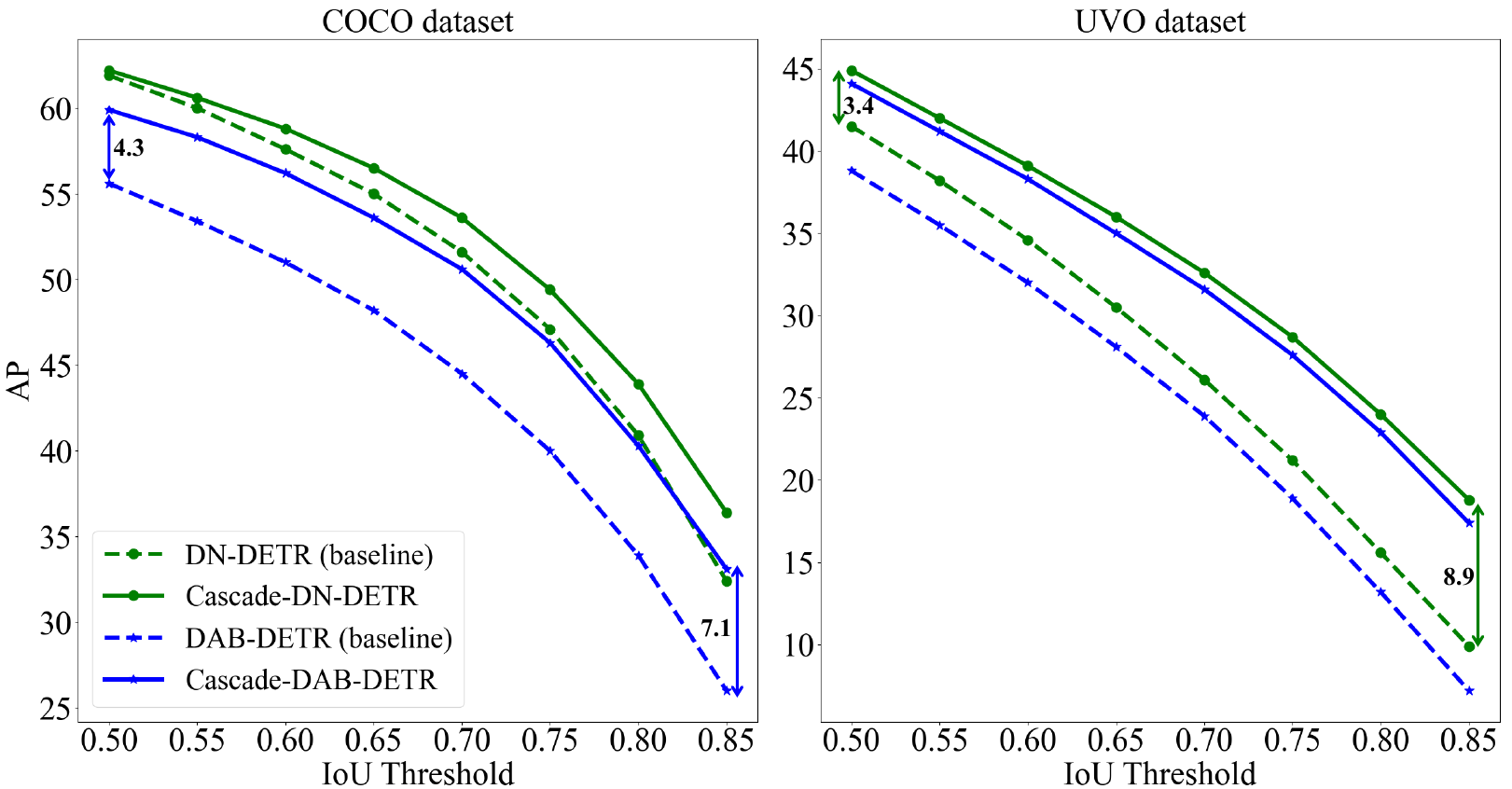}
\vspace{-0.35in}
\label{fig:strict_iou}
\end{figure}

Despite the notable progress of DETR-based detectors, there are still significant limitations that need to be addressed. Figure~\ref{fig:teaser} shows that DETR-based methods severely struggle when applied outside of the conventional COCO benchmarks. 
This can be attributed to the limited number of training samples and diverse styles encountered in more task-specific domains, resulting in a drop in performance even blow their CNN-based predecessors. In particular, we find that  on
e.g., Cityscapes~\cite{cityscapes} and Brain tumor~\cite{tumor} benchmarks, the performance of DN-DETR~\cite{dndetr} is substantially poorer than Faster R-CNN despite its superior performance on COCO.
Moreover, the prediction of highly accurate bounding boxes remains challenging. 
In Figure~\ref{fig:strict_iou}, given stricter IoU thresholds, existing DETR-based methods still have substantial room for improvement.

We partially attribute these two problems, namely, poor generalization to other datasets and limited bounding box accuracy, to a lack of a local object-centric prior. 
Following the general philosophy of transformers~\cite{dosovitskiy2020vit}, DETR-based methods replace convolutions with global cross-attention layers in the detection head, thus removing the object-centric inductive bias. 
We argue that without such bias makes it difficult to accurately identify local object regions, thus limiting the bounding box accuracy. 
Additionally, the reliance on a purely data-driven approach to learn such bias places a heavy reliance on large annotated datasets, which are often unavailable in diverse real-world applications.
Many detection tasks have distinct image domains, such as medical imaging or document analysis (as shown in Figure~\ref{fig:teaser}), which differ significantly from those in COCO or ImageNet, making pretraining on large annotated datasets even less effective.

The other attributing factor is the scoring of bounding box predictions which further exacerbates the high accuracy of DETR-based detectors. 
The query scoring in DETR decoder is purely based on the final classification confidence. However, these scores are largely oblivious of the quality of the predicted bounding box. 
Instead,  we argue that correctly classified box proposals that better overlaps with the ground-truth should be assigned higher scores.

To address these two issues, this paper presents \modelname to promote high-quality universal detection performance for DETR-based models. 
To tackle the lack of local object-centric prior, we introduce
cascade attention in the DETR decoder, which constrains the spatial cross attention layers to only inside the previously predicted bounding box of each query. Since DETR decoder has multiple decoder layers, the cascade structure iteratively refines the cross-attention region for each query, using a more accurate box prediction after every layer. 
To improve the scoring of box predictions, we propose an
IoU-aware Query Recalibration, by adding an IoU prediction branch to re-calibrate query scores. In parallel to the query classification and regression branches, the IoU prediction branch computes the box proposal IoU to the corresponding GT object. 
This enables each matched learnable query to be aware of its quality more accurately.
During inference, we recalibrate the classification scores by the predicted localization scores as the the final ones to rank proposals.

We further compose a new detection benchmark UDB10 and corresponding evaluation metric UniAP to support high-quality universal detection. We hope to facilitate the detection community not only focusing detection results on COCO but also in more wide real-life applications.
As in Table~\ref{tab:UDB10_1}, UDB10 consists of 10 datasets from various real-life domains. We compare the UniAP among Faster R-CNN~\cite{ren2015faster}, DN-DETR~\cite{dndetr} and Cascade-DN-DETR, where our approach achieves the best 44.2 UniAP. With negligible model parameters increase, our method significantly promotes the detection quality of DETR-based models for 5.7 UniAP, especially on the domain-specific datasets. This is also validated by our large performance gain in Figure~\ref{fig:strict_iou}.
On the large-scale COCO benchmark, Cascade-DN-DETR achieves significant 2.1 and 2.4 AP improvement over DN-DETR using R50 and R101 backbone respectively.

\begin{table}[!t]
\centering
\caption{Datasets components in UDB10 benchmark for evaluating high-quality universal object detection. The UniAP metric computes the mean of AP for each individual dataset component. Training is done individually on each dataset. All comparing methods use ResNet50 as backbone. Our Cascade-DN-DETR is built on  DN-DETR~\cite{dndetr}. FR-CNN: Faster R-CNN; Paintings: People in paintings dataset~\cite{painting}; Document: Document parts~\cite{document}.}
\vspace{0.05in}
\resizebox{1.0\linewidth}{!}{%
\begin{tabular}{l|cc|ccccccc}
\hline
 & Domain & \# Images & FR-CNN~\cite{ren2015faster} & DN-DETR~\cite{dndetr} & Ours & \\ \hline
COCO~\cite{lin2014microsoft}  & Natural   & 118k &  37.9 & 43.4 &  45.5$_{\uparrow\textbf{2.1}}$  \\ 
UVO~\cite{uvo}  &  Open World  & 15k & 24.7 &  22.3  &  28.4$_{\uparrow\textbf{6.1}}$  \\ 
Cityscapes~\cite{cityscapes}  &  Traffic  & 3k &  30.1 &  19.0  &  29.0$_{\uparrow\textbf{10.0}}$    \\ 
BDD100K~\cite{bdd100k}  &  Traffic  & 70k  & 31.0 & 28.2 & 30.2$_{\uparrow\textbf{2.0}}$     \\ 
Brain tumor~\cite{tumor} &  Medical  & 7k &  43.5 &  38.9  &  46.5$_{\uparrow\textbf{7.6}}$    \\ 
Document~\cite{document}  &  Office  & 1k & 48.0 &  44.7  & 50.9$_{\uparrow\textbf{6.2}}$  \\ 
Smoke~\cite{smoke} & Natural & 0.5k &  67.1 &  66.5 & 71.8$_{\uparrow\textbf{5.3}}$ \\
EgoHands~\cite{egohands} &  Egoview  & 11k & 74.9 &  74.4  & 77.6$_{\uparrow\textbf{3.2}}$   \\
PlantDoc~\cite{plantdoc}   &  Natural  & 2k & 38.9 & 45.0  & 49.1$_{\uparrow\textbf{4.1}}$ \\
Paintings~\cite{painting}   & Art   & 0.6k & 17.0 &  2.2  &   13.4$_{\uparrow\textbf{11.2}}$ \\
\hline
\textbf{UniAP}  &    &  & 41.3 &  38.5  & \textbf{44.2}$_{\uparrow\textbf{5.7}}$\\
\hline
\end{tabular}}
\label{tab:UDB10_1}
\vspace{-0.25in}
\end{table}

%------------------------------------------------------------------------
\section{Related Work}
\parsection{DETR-based Object Detection}
Modern object detectors can be mainly divided into the classical CNN-based and more recent DETR-based models~\cite{chen2022recurrent,zhang2022accelerating,li2022exploring,cao2022cf, wang2021pnp}. The convolutional detectors includes one-stage detectors~\cite{yolov3, tian2020fcos} and two/multi-stage models~\cite{ren2015faster,girshick2015fast, cai2018cascade,chen2019hybrid}. 
For DETR-based models~\cite{carion2020detr, zhu2020deformable, meng2021conditional, updetr, roh2022sparse}, recent works such as~\cite{dabdetr, dndetr, dino} outperform CNN-based detectors by a significant margin on COCO.

For improving the transformer decoder, Dynamic DETR~\cite{dai2021dynamic} designs dynamic encoder for focusing on more important features on multi-scale feature maps while~\cite{sun2021rethinking} even replaces the decoder with FCOS/RCNN networks.
To enhance decoder queries, Efficient DETR~\cite{yao2021efficient} adopts the top-K locations from encoder’s dense prediction prior. 
Anchor DETR~\cite{wang2022anchor} represents object queries based on anchor points, while DAB-DETR~\cite{dabdetr} adopts 4D anchor box coordinates. 
DN-DETR~\cite{dndetr} further speeds up the DETR converge by an additional denoising branch.
Based on DN-DETR and DAB-DETR, DINO~\cite{dino} includes contrastive denoising training and mixed query selection for anchor initialization.

In contrast to existing DETR-based methods~\cite{gao2021fast, sun2021rethinking,liu2021wb}, \modelname is targeted for high-quality object detection. 
The proposed cascade attention and IoU-aware query recalibration significantly improves AP performance under strict IoU thresholds.
Besides only experimenting on COCO, we show the effectiveness of our approach on the constructed UDB10 benchmark, which contains a wide range of task-specific applications.

\parsection{Cross-attention in DETR-based Decoder} 
In addition to standard cross-attention~\cite{carion2020detr} applied on global image features, Deformable DETR~\cite{zhu2020deformable} proposes deformable attention. A set of 2D image locations are predicted, which are then used for attention.
Mask2Former~\cite{cheng2022masked} proposes mask attention, only indented for segmentation. 
Different from these methods, our cascade attention utilizes the iteratively updated boxes to constrain the cross-attention on the image, and does not introduce any extra model parameters. We reveal our advantages to deformable attention and mask attention in the experiment section.

\parsection{High-quality Object Detection}
Different from high-quality segmentation networks~\cite{sam_hq,cheng2022masked,transfiner,vmt} based on transformers, existing works~\cite{szegedy2014scalable,cai2018cascade,cao2020d2det, chen2019hybrid} on high-quality object detection are mainly R-CNN based. Sepcially, Cascade R-CNN~\cite{cai2018cascade} introduces multi-stage detectors trained with increasing IoU thresholds, while Dynamic R-CNN~\cite{zhang2020dynamic} designs dynamic labels and a regression loss. Wang et al.~\cite{huang2019mask} improves R-CNN based segmentation via mask scoring. For localization quality estimation (LQE), previous works~\cite{tian2020fcos,jiang2018acquisition,zhu2019iouuniform,chen2021disentangle,li2020generalized} mainly study it in FCOS or R-CNN based detectors.
To our knowledge, we are the first DETR-based method tackling the problem of predicting highly accurate boxes.

\parsection{DETR-based Universal Object Detection} Existing DETR-based methods~\cite{dndetr,dino} mostly train and evaluate their performance on COCO. However, detectors should generalize well to wide and practical scenarios, such as medical imaging and document analysis. 
Typically these datasets contain around 1K to 20K images, where some contain images  with very different styles than COCO/ImageNet.
Different from previous detection works on adaptation learning~\cite{wang2019towards,chen2018domain}, few-shot setting~\cite{zhang2022meta} or mixed training~\cite{zhou2022simple}, we focus on the fully supervised training setting per dataset to evaluate the detector performance in various application scenarios. To facilitate the research on universal object detection using DETR-based detectors, we construct a large-scale UDB10 benchmark containing 228k images, which is doubles the size of UODB~\cite{wang2019towards} for domain adaptation and has significantly more images per dataset component than~\cite{ciaglia2022roboflow}.
We show that \modelname with injected local object-centric prior brings large performance gains to existing DETR-based models across wide and challenging domains, making DETR-based models more universally applicable.

%------------------------------------------------------------------------
\section{\modelname}
\begin{figure}[!t]
\centering
\includegraphics[width=1.0\linewidth]{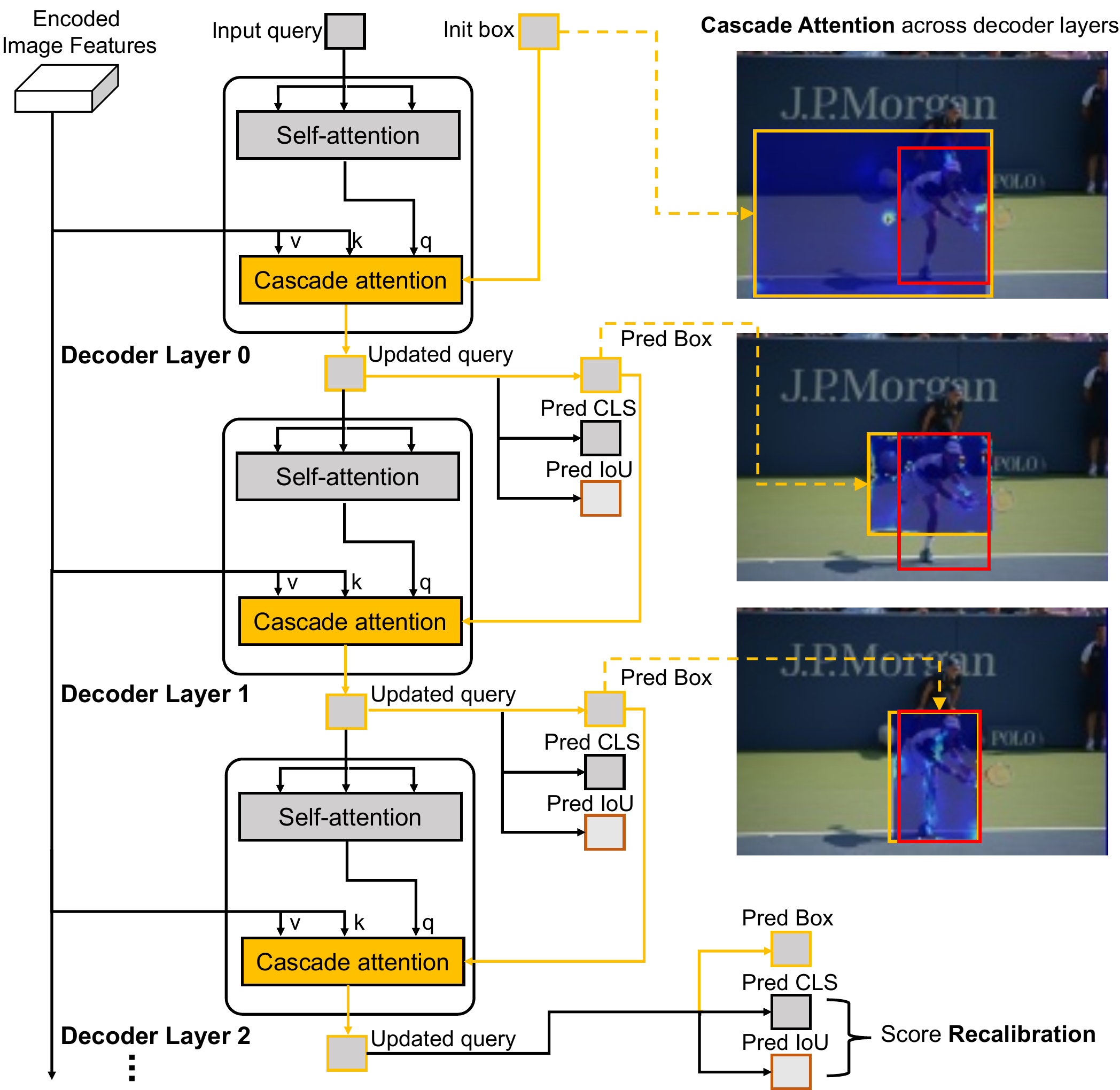}
\vspace{-0.1in}
\caption{The transformer decoder of our Cascade DETR. We feed in the encoded image features from the transformer encoder along with learnable queries. The box-constrained cross-attention regions (inside the yellow predicted boxes) are iteratively refined per decoder layer, which, in turn, further promotes the detection accuracy. The score recalibration is used in the last transformer decoder layer during inference. Red box denotes the ground truth object box. We omit the transformer encoder and positional embedding for clarity.}
\label{fig:architectural}
\end{figure}

We propose \modelname for high-quality and universal object detection. We first review the design of conventional DETR decoder in Section~\ref{sec: convention_detr}. Then we introduce our detection transformer \modelname in Figure~\ref{fig:architectural}. It is an iterative approach consisting of 
two novel components: \textbf{1)} Cascade attention, which constrains the cross-attention range in each decoder layer within the box region predicted from the preceding layer (Section~\ref{sec: box_att}); \textbf{2)} Query-recalibration, which recalibrates the learnable queries with the IoU prediction to enable more accurate query scoring (Section~\ref{sec: query_recab}).
Finally, we describe the training and inference details of our \modelname in Section~\ref{sec:details}.

\subsection{Preliminaries: The DETR Decoder}
\label{sec: convention_detr}

We briefly review the design of the standard DETR decoder, which
consists of a set of cross- and self-attention layers that iteratively updates a set of queries, initialized as learnable constants. At the $i$-th layer, the queries $\textbf{Q}_i \in \mathbb{R}^{N\times D}$ are first input to a self-attention block, followed by cross-attention with the encoded image features of size $H\times W\times D$.
The cross-attention is computed as the weighted sum over the global feature map,
\begin{equation}
	\label{eq:attention}
	\textbf{Q}_{i+1} = \sum_{j=1}^{H\times W}\frac{\exp(f_{q}(\textbf{Q}_{i})\cdot \textbf{K}_{i}^{j})\textbf{V}_{i}^j}{\sum_{k}\exp(f_q(\textbf{Q}_{i})\cdot \textbf{K}_{i}^{k})} + \textbf{Q}_{i}\,,
\end{equation}
where $\textbf{K}$ and $\textbf{V}$ respectively denote key and value maps extracted from the image features. The index $i$ denotes the cross-attention layer, $j$ is the 2D spatial location on the image, and $f_q$ denotes the query transformation function.

The updated queries $\textbf{Q}_{i+1}$ are then used to predict bounding boxes $\textbf{B}_{(i+1)}$ and query scores $\textbf{S}_{(i+1)}$ by feeding them into two parallel linear layers $f_\text{box}$ and $f_\text{score}$ respectively,~\ie, $\textbf{B}_{(i+1)} = f_\text{box}(\textbf{Q}_{(i+1)})$ and $\textbf{S}_{(i+1)} = f_\text{cls}(\textbf{Q}_{(i+1)})$.
The query score matrix $\textbf{S}_{(i+1)}$ of size $N \times (C+1)$ contains the class probabilities for all input queries, where $C$ is the number of classes of the dataset. This decoder design is generally used in~\cite{carion2020detr,dabdetr,meng2021conditional,dndetr}.

\subsection{\modelname Architecture} 
In this section, we describe the architecture of~\modelname, which injects local object-centric bias into the conventional transformer decoder in Section~\ref{sec: convention_detr}.
Similar to existing DETR-based methods, such as DAB-DETR~\cite{dabdetr} and DN-DETR~\cite{dndetr}, our architecture contains a transformer encoder for extracting image features. 
The encoded features combined with the positional encoding are fed to the transformer decoder.
The learnable queries are also fed into the decoder to localize and classify objects through cross-attention.
The two new modules in our \modelname are cascade attention and IoU-aware query re-calibration, which only bring negligible computation overhead or model parameters while significantly improving the detection quality and generalizability.

\subsection{Cascade Attention}
\label{sec: box_att}
\begin{figure}
\includegraphics[width=1.0\linewidth]{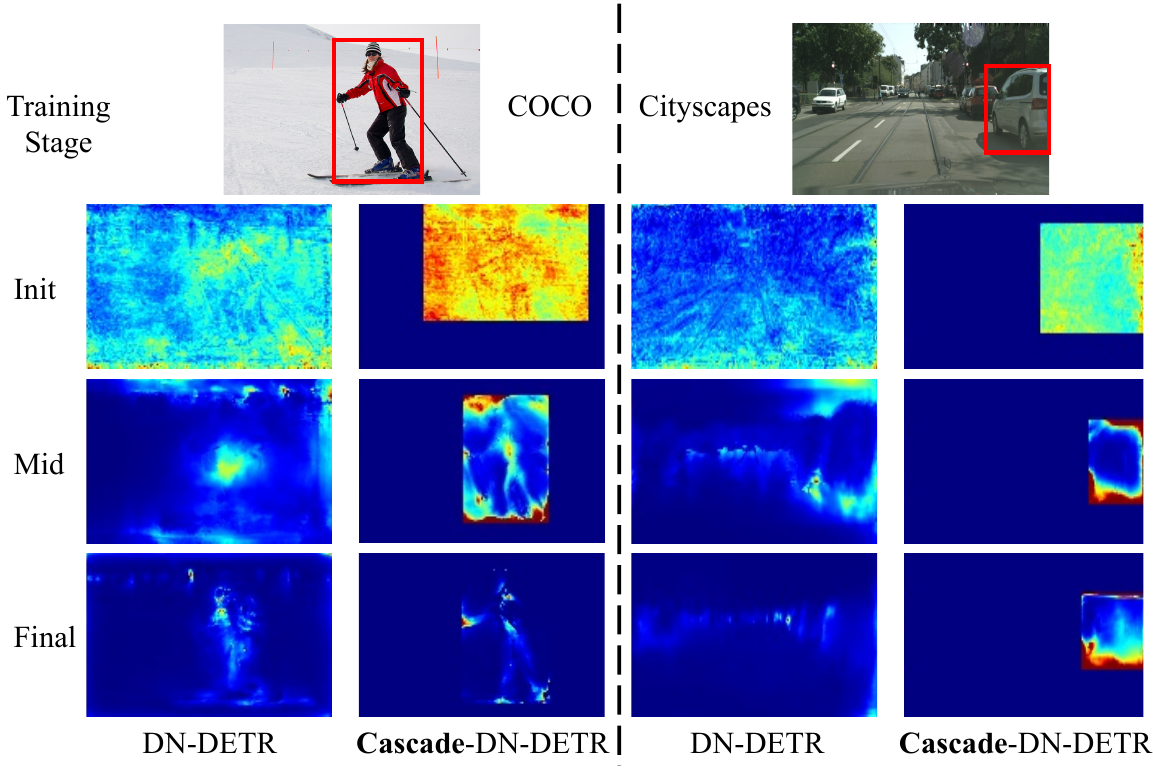}
\caption{Visual comparison of cross-attention map between DN-DETR~\cite{dndetr} and our Cascade-DN-DETR on COCO and Cityscapes datasets. At different network training stages, we visualize the cross-attention map of the last transformer decoder layer, where the learnable query corresponds to the object inside the red box.}
\label{fig:Cascade Attention}
\end{figure}

In the standard DETR decoder, learnable queries attend globally over the entire image features, as in Eq.~\ref{eq:attention}. 
However, to accurately classify and localize the object, we argue that local information around each itself object is most crucial. The global context can be extracted via self-attention between queries. 
In Figure~\ref{fig:Cascade Attention}, we observe that the cross-attention distribution during COCO training tends to converge to the surrounding regions of the predicted object locations.
While transformer model can learn this inductive bias end-to-end, it requires large amounts of data. This problem becomes more pronounced for small or task-specific datasets with image styles radically different from those exhibited in ImageNet.

To address the above issue, we treat the object-centric prior as a known constraint to incorporate into both the initialization and training procedures, as depicted in the Figure \ref{fig:architectural}. 
We design the cascade attention in layer $i+1$ as,
\begin{align}
	\label{eq:box_attention}
	\textbf{Q}_{i+1} &= \frac{1}{Z_i}\sum_{j \in \textbf{S}_i}\frac{\exp(f_{q}(\textbf{Q}_{i})\cdot \textbf{K}_{i}^{j})\textbf{V}_{i}^j}{\sum_{k \in \textbf{S}_i}\exp(f_q(\textbf{Q}_{i})\cdot \textbf{K}_{i}^{k})} + \textbf{Q}_{i}\,, \\
    \textbf{S}_{i} &= \mathcal{M}(\textbf{B}_{i}) = \mathcal{M}(f_\text{box}(\textbf{Q}_{i}))\,,
\end{align}
where ${\bf S}_i$ is the set of 2D locations inside the predicted bounding box $\textbf{B}_{i}$ from the preceding decoder layer $i$. 
The cascade structure utilizes the property that the predicted $\textbf{B}_{i}$ will be more accurate after every decoder layer in DETR-based detectors~\cite{carion2020detr}. 
Thus, the box-constrained cross-attention region~${\bf S}_i$ not only brings object-centric bias, but will also be iteratively refined (see Figure~\ref{fig:architectural}).
With more accurately cross-attended features per layer, cascade attention in turn promotes the detection accuracy per layer.

We validate our assumption by visualizing the attention map in Figure \ref{fig:Cascade Attention}. The initial and final attention maps of a baseline DN-DETR model are shown both in COCO and Cityscapes. On COCO, we observe both the cross-attention of a randomly initialized query eventually converges on semantically distinct locations using DN-DETR or Cascade-DN-DETR. However, on Cityscapes, there is a obvious contrast between the two methods, where the integration of object-centric knowledge is more important to focus the attention on the most relevant parts of the image.

Unlike previous approaches such as DAB-DETR~\cite{dabdetr} and Deformable DETR~\cite{zhu2020deformable}, which utilize soft constraints, the design of our \modelname is much simpler. The prediction boxes in each layer of the DETR decoder is directly used as constraints to limit the cross-attention range in the following layer. This inductive bias enables DETR to converge quickly and achieve superior performance, especially for small and diverse datasets.

\subsection{IoU-aware Query Recalibration}
\label{sec: query_recab}
Most DETR-based detectors take 300~\cite{dndetr,dabdetr} or even 900~\cite{dino} learnable queries as input to the transformer decoder and predicts one box per query.
When computing final detection results, classification confidence is adopted as a surrogate to rank all query proposals. However, the classification score does not explicitly account for the accuracy of the predicted bounding box, which is crucial for selecting high-quality proposals. We therefore introduce an IoU-aware scoring of the predicted queries in order to achieve a more well-calibrated confidence, which better reflects the quality of the predictions.

Instead of scoring queries by the classification confidence, we score them by the expected IoU with the ground-truth box. Let $E(\text{IoU}_q)$ be the expected ground-truth IoU of query $q$. Further, let $P(\text{obj}_q)$ denotes the probability of $q$ indicating an object, as obtained from the classification probability. The expected IoU of a query is computed as 
\begin{align}
\label{eq:recalibra}
&E(\text{IoU}_q)  \nonumber \\
&= E(\text{IoU}_q \mid\text{obj}_q) P(\text{obj}_q) + E(\text{IoU}_q \mid\neg \text{obj}_q) P(\neg \text{obj}_q) \nonumber \\
&= E(\text{IoU}_q \mid\text{obj}_q) P(\text{obj}_q)
\end{align}
Here, $\neg$ denotes negation of the binary random variable.
The second equality follows from that the expected IoU for a prediction that is not an object is zero: $E(\text{IoU}_q \mid\neg \text{obj}_q)=0$. 

To predict the expected IoU \eqref{eq:recalibra}, we introduce an additional branch that predicts the expected IoU for a present ground-truth object $E(\text{IoU}_q \mid\text{obj}_q)$, as illustrated in Figure~\ref{fig:architectural}. Specifically, we simply use another linear layer in parallel to the classification and box regression branches. As derived in Eq.\eqref{eq:recalibra}, the final query score is then obtained as the product between the predicted IoU and the original classification confidence $P(\text{obj}_q)$.

We supervise the IoU prediction with an $L_2$ loss to the ground-truth IoU, denoted $\text{IoU}_q^\text{GT}$,
\begin{equation}
    \label{eq:ioutrain}
    L_\text{IoU} = \left\| E(\text{IoU}_q \mid\text{obj}_q) - \text{IoU}_q^\text{GT} \right\|^2 \,.
\end{equation}
The loss is only applied for queries $q$ with an assigned ground-truth, as we condition on the presence of the object in the expectation.
Note that the $L_2$ loss implies learning the mean, \ie expectation, of a Gaussian distribution over the IoU values. We ablate this choice of loss in Table~\ref{tab:iou_losstype} of the experiment section.

\begin{figure}[!t]
\centering
\includegraphics[width=0.9\linewidth]{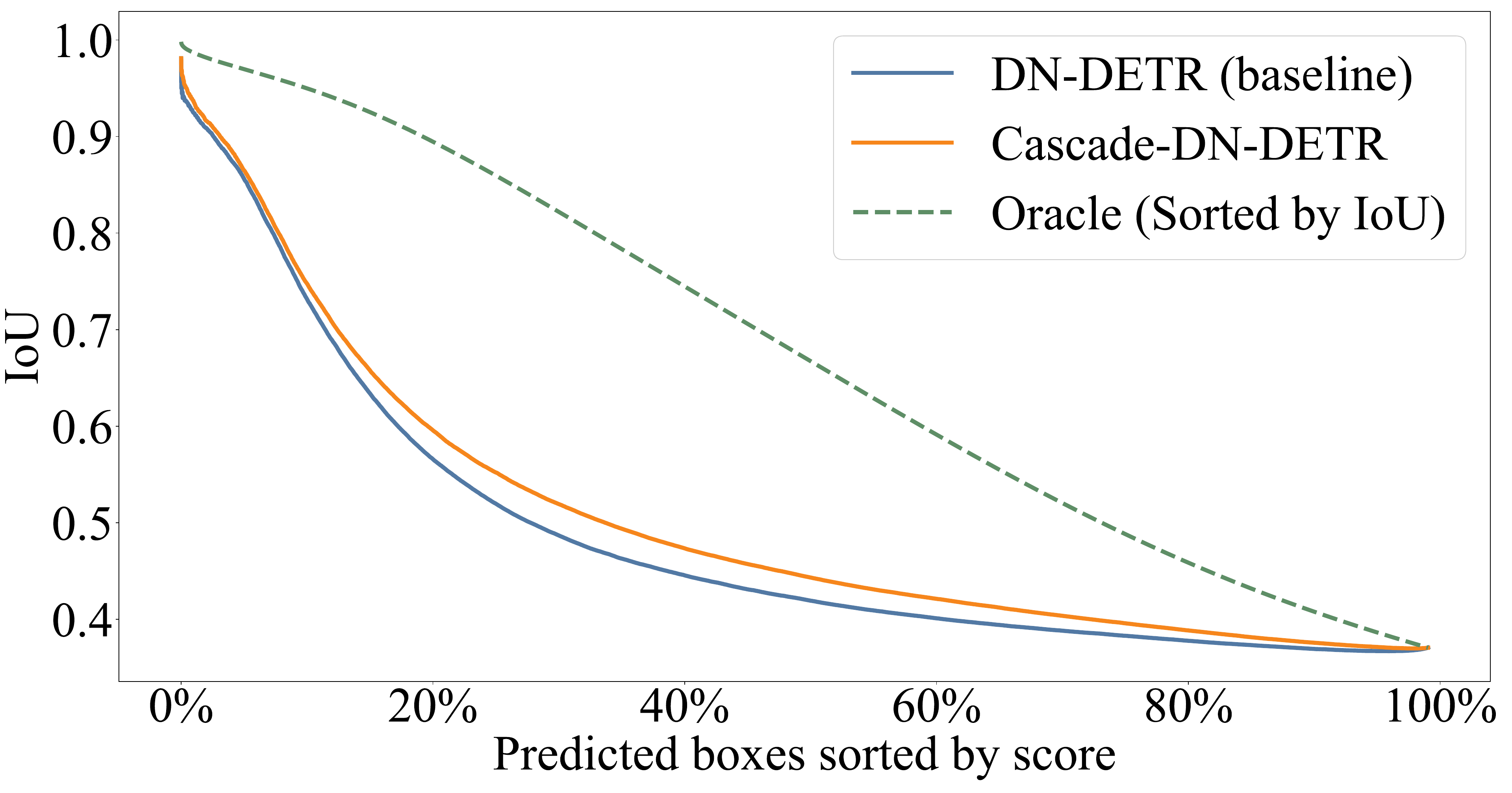}
% \vspace{-0.25in}
\caption{Sparsification plot between query localization quality (IoU to GT boxes) and query ranking (scoring). For 5k COCO validation images with 50 outputs for each image, we sort all the outputs by their confidence scores. We then compute the IoU with ground truth for each prediction and show a cumulative average of IoU. \textbf{Oracle: } Cumulative average of IoU sorted by IoU itself. Compared to the blue curve before recalibration, ours re-calibrated orange curve is closer to the Oracle and has a much higher localization quality. }
\label{fig:distribution}
\vspace{-0.2in}
\end{figure}

To analyze the advantage of our IoU-aware query recalibration, we generate sparsification plots over all predictions on COCO in Figure~\ref{fig:distribution}. All predictions are sorted with respect to the confidence score. The average IoU with ground-truth is then plotted for the $N$ predictions with the highest confidence score, by varying $N$ across the x-axis. The Oracle represents the upper bound, obtained by taking the top $N$ predictions in terms of ground-truth IoU. Comparing to Cascade-DN-DETR without query recalibration (blue curve), our recalibrated result (orange curve) achieve substantially better ranking of the results, leading to a higher IoU.

\subsection{Training and Inference}
\label{sec:details}

Our \modelname is
trained in an end-to-end manner using a multi-task loss function,
\begin{equation}
\mathcal{L}_{\text{Detect}} = \mathcal{L}_{\text{Box}} + \lambda_1\mathcal{L}_{\text{Class}} + \lambda_2\mathcal{L}_{\text{IoU}},
\end{equation}
where~$\mathcal{L}_{\text{Detect}}$ supervises both the position prediction and the category classification borrowed from the DETR~\cite{carion2020detr} detector.
The hyper-parameters $\lambda_1$ and $\lambda_2$ balances the loss functions, and set to $\{1.0, 2.0\}$ respectively on the validation set.
Following~\cite{dndetr,dabdetr}, FFNs and the Hungarian loss are adopted after each decoder layer. FFNs share their model parameters in each prediction layer.

During inference, our cascade attention is consistently used as it only relies on the predicted boxes in each transformer decoder layer. 
For the query scoring calibration manner, as described in~\ref{eq:recalibra}, we only apply it on the final transformer decoder layer. 

%------------------------------------------------------------------------
\section{Experiments}\label{sec:exp}
\subsection{Experimental Setup}

\parsection{COCO} We perform evaluation on the challenging MS COCO 2017 object detection benchmark~\cite{lin2014microsoft}. Models are trained with 118k training images in \textit{train2017} split and evaluated on the 5k validation images in \textit{val2017}.
We report the standard average precision (AP) result under different IoU thresholds. 

\parsection{UVO and Cityscapes} To generalize on universal object detection, we also conduct experiments on two challenging datasets, UVO~\cite{uvo} and Cityscapes~\cite{cityscapes}. UVO is an exhaustively labeled open-world dataset with 15k training images and 7k validation images. Cityscapes is an urban street scene dataset which contains 3k training images and 500 validation images. We perform results comparison following the standard training and model evaluation setting on the two benchmarks.

\parsection{UDB10 Benchmark} There is a wide variety of detection applications in real-life scenarios. To facilitate the research on universal detection, we construct a large-scale UDB10 benchmark which is composed of 10 different datasets across wide domains.
Besides the aforementioned COCO~\cite{lin2014microsoft}, Cityscapes~\cite{cityscapes} and UVO~\cite{uvo}, the other 7 task-specific datasets includes BDD100K~\cite{bdd100k}, Brain Tumor~\cite{tumor}, Document Parts~\cite{document}, Smoke~\cite{smoke}, EgoHands~\cite{egohands}, PlantDoc~\cite{plantdoc} and People in paintings~\cite{painting}. 
UDB10 contains 228k images, which covers a great variety of domains such as medical, traffic, nature, office, art, ego-view, etc. 
We follow the official training/evaluation settings on each dataset component.
Along with the UDB10 benchmark, we design UniAP metric to evaluate the %high-quality 
detection performance among detectors. After detectors are trained individually on each dataset component, UniAP is computed as the mean over the AP scores across all datasets. 

In Table~\ref{tab:data_comp}, we compare UDB10 with two other existing universal detection benchmarks UODB~\cite{wang2019towards} and Roboflow 100~\cite{ciaglia2022roboflow}, where we find UDB10 has significantly more images and annotated instances per dataset component. We establish UDB10 aims to evaluate the detection performance of data-senstive DETR-based methods in diverse domains.

\parsection{Implementation Details} In our experiments, we use two different backbones: ResNet-50 and ResNet-101 pre-trained on ImageNet-1k, and train our model with an initial learning rate  $1\times10^{-5}$ for backbone and $1\times10^{-4}$ for transformer. We use the AdamW optimizer with weight decay   $1\times10^{-4}$.  We train on 8 Nvidia GeForce RTX 3090 GPUs with total batch size of 8, and adopt two training schedules. For small datasets (less than 10k images), we train DETR-based methods for 50 epochs with a learning rate decay after 40 epochs.  
For large datasets (greater than or equal to 10k images), we adopt DETR-based methods for 12 epochs with a learning rate decay after 10 epochs.
The original DETR uses 100 queries, while in all other experiments we use 300 queries except DINO~\cite{dino}, where 900 queries are used to be consistent with their paper. For multi-scale features, we use DN-Deformable-DETR~\cite{zhu2020deformable} with a deformable encoder. For the first layer cascade attention input box, we use the initial learnable anchor box proposed in DAB-DETR~\cite{dabdetr}. For Faster-RCNN, we use 1X schedule for large datasets and 3X schedule for small datasets. For mask attention ablation on COCO, we train an extra mask head with ground truth mask annotations and do not use query recalibration. More details are in the Supp.\ file.

\begin{table}[!t]
\centering
\caption{Comparison between the universal detection benchmarks. \# Images / set denotes the average number of training images per dataset component, while \# Instances / set is the average number of annotated boxes per dataset component.}
\vspace{0.05in}
\resizebox{0.95\linewidth}{!}{%
\begin{tabular}{l|c|c|c}
\hline
Benchmarks   &  \# Images &  \# Images / set  & \# Instances / set   \\ \hline
UODB~\cite{wang2019towards} & 113k   &  10.2k &  69k \\
Roboflow 100~\cite{ciaglia2022roboflow} &  224k & 2.2k  &  25k   \\ \hline
UDB10 (Ours) & 228k & 22.8k & 239k    \\ \hline
\end{tabular}}
\label{tab:data_comp}
\vspace{-0.25in}
\end{table}

\subsection{Ablation Study}
We conduct detailed ablation studies for \modelname using ResNet-50 as backbone on the Cityscapes~\cite{cityscapes} and UVO~\cite{uvo} datasets.
We analyze the impact of each proposed component of our \modelname.

\parsection{Ablation on Cascade Attention (CA)} In Table~\ref{tab1}, we study the effect of Cascade Attention (CA). Built on the baseline DN-DETR, CA significantly promotes the performance for 3.7 AP on UVO and 9.9 AP on Cityscapes. In Table~\ref{tab:att_comp}, we further compare our cascade attention in the transformer decoder to the mask attention~\cite{cheng2022masked}. We perform comparisons on both UVO and COCO as both these two datasets in UDB10 have corresponding GT mask labels per box. We design the mask attention by an additional mask prediction branch, which is supervised by the GT mask labels. This can be regarded as an oracle analysis as many object detection benchmarks have no annotated GT masks labels. Our cascade attention achieves similar results to mask attention by improving 0.6 AP on COCO but decreasing 0.6 AP on UVO. This indicates that accurate object mask shape is not necessary for object detection.

\parsection{Ablation on Query Recalibration (QR)} 
In Table~\ref{tab1}, we also validate the effect of Query Recalibration (QR), which promotes 3.6 AP on UVO and 4.1 AP on Cityscapes. Specifically, on UVO, QR improves 4.9 AP$_{75}$ which is much larger than gain of 3.0 AP$_{50}$. 
We further perform detailed ablation experiments on the query recalibration loss types (Table~\ref{tab:iou_losstype}), recalibration methods (Table~\ref{tab:iou_manner}) and recalibration training strategies (Table~\ref{tab:iou_strategy}). In Table~\ref{tab:iou_losstype}, the performance boost is similar using L2 or L1 loss while outperforming Huber Loss with 0.6 AP. 

As derived in Eq.~\ref{eq:recalibra}, our expected IoU is computed as a product between the classification confidence and IoU prediction. Table~\ref{tab:iou_manner} compares this fusion with other strategies for computing the final query score. Our principled approach achieves the the best performance of 25.9 AP. It outperforms the baseline classification-only by 3.6 AP and the sum fusion by a large margin of 2.4 AP. We also compare with directly predicting a single confidence score, supervised both by the baseline classification loss and our IoU loss (second row). While achieving a significant gain of 2.5 AP over the baseline, it does not reach the performance of our derived expected IoU based fusion.

Since the expected IoU scores in Eq.~\ref{eq:ioutrain} are conditioned on the presence of the object, we only add this loss on predictions which are matched with ground-truth boxes. We ablate this choice in Table~\ref{tab:iou_strategy} by adding the loss to all predictions. The latter results in a performance only marginally above the baseline without IoU-awareness. Again, this demonstrate the advantage of our principled IoU-based query scoring.

\subsection{Comparison with State-of-the-art}

We compare \modelname with the state-of-the-art object detection methods on COCO, UVO, Cityscapes and our constructed UDB10 benchmark. We integrate \modelname on three representative methods~\cite{dndetr,dabdetr,dino}, and find that \modelname attains consistent large gains over the strong baselines.

\begin{table}[t!]
\centering
\caption{Ablation study on the Cascade Attention (CA) and Query Recalibration (QR). We use ResNet-50 based DN-DETR~\cite{dndetr} with deformable encoder as our baseline.}
\vspace{0.05in}
\resizebox{1.0\linewidth}{!}{%
\begin{tabular}{l|cc|cccc|ccc}
\hline
   \multicolumn{2}{c}{} & & \multicolumn{4}{c|}{UVO} & \multicolumn{3}{c}{Cityscapes} \\ \hline
Model  & CA & QR & \multicolumn{1}{l}{$AP$} & \multicolumn{1}{l}{$AP_{50}$} & \multicolumn{1}{l}{$AP_{75}$} & $AR$ & $AP$ & \multicolumn{1}{l}{$AP_{50}$} & \multicolumn{1}{l}{$AP_{75}$} \\ \hline
DN-DETR~\cite{dndetr} &    &    &  22.3  &  41.5   &   21.2  &  51.4  & 19.0  &   39.3   &   15.7 \\
&   $\checkmark$  &   & 26.0$_{\uparrow\textbf{3.7}}$ &  43.0  &  25.6  &  57.8  & 28.9$_{\uparrow\textbf{9.9}}$    & 52.1    &  27.0  \\
&   &   $\checkmark$  &  25.9$_{\uparrow\textbf{3.6}}$ &  44.5  &  26.1  &  53.9  &   23.1$_{\uparrow\textbf{4.1}}$   &   44.1   &  20.7  \\
Ours    & $\checkmark$  & $\checkmark$      &  28.4$_{\uparrow\textbf{6.1}}$ &  44.9  & 28.7  &   58.2   &  29.0$_{\uparrow\textbf{10.0}}$    &  49.5 & 28.4  \\ \hline
\end{tabular}}
\label{tab1}
\end{table}

\begin{table}[t!]
\centering
\caption{Detection performance comparison between cascade and mask cross-attention schemes in the transformer decoder on UVO and COCO. Both two cross-attention schemes are taking DN-Deformable-DETR as baseline. \textbf{Oracle:} We add an extra mask head with GT mask supervision, and use predicted outputs as attention mask in the transformer decoder.}
\vspace{0.05in}
\resizebox{1.0\linewidth}{!}{%
\begin{tabular}{l|cccc|ccc}
\hline
   \multirow{2}{*}{Cross-attention Type} & \multicolumn{4}{c|}{UVO} & \multicolumn{3}{c}{COCO} \\ \cline{2-8}
  & $AP$ & $AP_{50}$ & $AP_{75}$ & $AR$ & $AP$ & $AP_{50}$ & $AP_{75}$ \\ \hline
Mask Attention (\textbf{Oracle}) &   29.0 & 46.4  &  29.5  &  56.8 &  44.2 & 61.1 & 47.8  \\
Cascade Attention (\textbf{Ours}) &  28.4$_{\downarrow\textbf{0.6}}$  & 44.9   &  28.7 & 58.2 &  44.8$_{\uparrow\textbf{0.6}}$ &  62.7 & 48.4  \\ \hline
\end{tabular}}
\label{tab:att_comp}
\end{table}

\begin{table}[t]
\centering
\caption{Ablation study on the query recalibration loss on the UVO dataset. \textbf{Baseline:} DN-Deformable-DETR.}
\vspace{0.05in}
\resizebox{0.75\linewidth}{!}{%
\begin{tabular}{l|cccc}
\hline
Loss type   & \multicolumn{1}{l}{$AP$} & \multicolumn{1}{l}{$AP_{50}$} & \multicolumn{1}{l}{$AP_{75}$} & $AR$ \\ \hline
Baseline &   22.3  &  41.5   &   21.2  &  51.4  \\ \hline
Huber Loss & 25.3$_{\uparrow\textbf{3.0}}$ &43.8  &  25.4  &  53.6 \\ 
L1 Loss &  25.9$_{\uparrow\textbf{3.6}}$ &  44.6  &  26.3  &  53.9    \\ 
L2 Loss &  25.9$_{\uparrow\textbf{3.6}}$ &  44.5  &  26.1  &  53.9    \\ \hline
\end{tabular}}
\label{tab:iou_losstype}
\vspace{-0.1in}
\end{table}

\begin{figure*}[!t]
\centering
\includegraphics[width=1.0\linewidth]{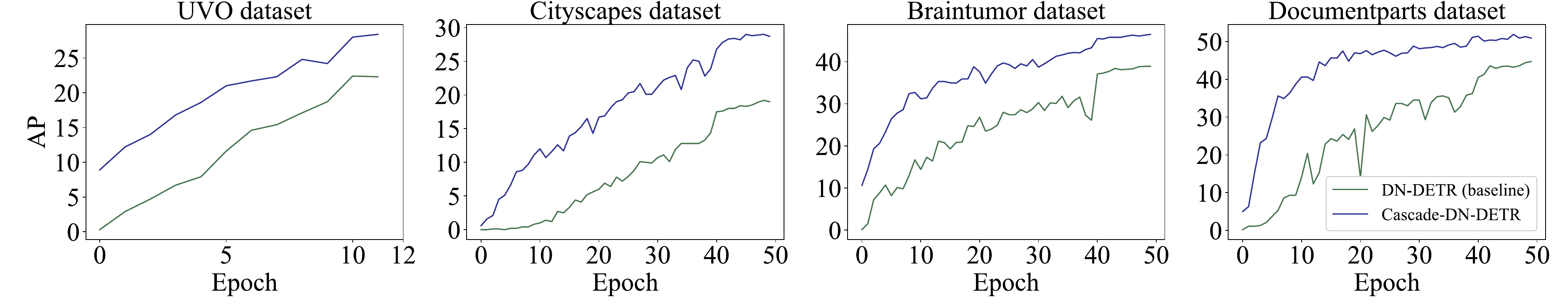}
\vspace{-0.2in}
\caption{Detection results comparison between DN-DETR~\cite{dndetr} (Baseline) and Cascade-DN-DETR (Ours) per training epoch on UVO~\cite{uvo}, Cityscapes~\cite{cityscapes}, Brain tumor~\cite{tumor}, and Documentparts~\cite{document} datasets. These datasets cover four various detection application domains. Cascade-DN-DETR achieves stable performance growth during training, and consistently outperforms the strong baseline DN-DETR~\cite{dndetr} with a significant margin. Note that DN-DETR has already been significantly sped up during training by its denoising branch.}
\vspace{-0.1in}
\label{fig:converge_speed}
\end{figure*}

\parsection{COCO} 
Table~\ref{tab:coco} compares~\modelname with state-of-the-art object detection methods on COCO benchmark. By integrating with SOTA DETR-based detectors, \modelname achieves consistent improvement on different backbones with negligible increase in model parameters, demonstrating its effectiveness by outperforming DN-Def-DETR~\cite{dndetr} by 2.1 AP and 2.4 AP respectively on R50 and R101 backbone. 
\modelname consistently attains larger increase in the strict AP$_{75}$ than the loose AP$_{50}$, which reveals our advantages in predicted box quality.
Using R50 as backbone, we also compare Cascade-DINO to DINO~\cite{dino} by replacing its deformable attention~\cite{zhu2020deformable} in the transformer decoder by our cascade attention. Cascade-DINO outperforms DINO by 1.0 AP$_{75}$ with a much simpler attention design, removing the necessity for predicting 2D anchor points and sampling offsets. 

\parsection{UVO and Cityscapes} 
Table~\ref{tab:uvo} tabulates the results on UVO benchmark, and Table~\ref{tab:cityscapes} tabulates the results on Cityscapes benchmark. \modelname achieves the best 28.4 AP on UVO, where
our approach significantly surpasses the strong baselines DN-DETR~\cite{dndetr} and DAB-DETR~\cite{dabdetr}, respectively with a large margin of 8.7 and 7.5 points in AP$_{75}$.
The significant increase in AP$_{75}$ is also consistent on Cityscapes. Comparing to our baseline DN-DETR, in Table~\ref{tab:cityscapes}, Cascade-DN-DETR substantially improves the AP$_{75}$ from 15.7 to 28.4.

\begin{table}[t]
\centering
\caption{Comparison on various score recalibration methods between classification (cls) score and predicted IoUs on UVO dataset during testing. \textbf{Baseline:} DN-Deformable-DETR trained without query recalibration.}
\vspace{0.05in}
\resizebox{1.0\linewidth}{!}{%
\begin{tabular}{l|cccc}
\hline
Scoring Manner   & \multicolumn{1}{l}{$AP$} & \multicolumn{1}{l}{$AP_{50}$} & \multicolumn{1}{l}{$AP_{75}$} & $AR$ \\ \hline
Baseline &   22.3  &  41.5   &   21.2  &  51.4  \\ \hline
Single score (cls.\ \& IoU superv.) & 24.8$_{\uparrow\textbf{2.5}}$ & 44.3  & 24.4 & 53.7    \\
Sum Fusion (cls.\ prob $+$ IoU) & 23.5$_{\uparrow\textbf{1.2}}$ &38.5 & 24.4 & 51.4    \\
Expected IoU (cls.\ prob $\times$ IoU) & 25.9$_{\uparrow\textbf{3.6}}$ &44.5 & 26.1 &  53.9 \\ \hline
\end{tabular}}
\label{tab:iou_manner}
\end{table}

\begin{table}[!t]
\centering
\caption{Ablation study on the training strategies for query recalibration on UVO. \textbf{Baseline:} Default DN-Deformable-DETR training manner.~\textbf{All:} Input all queries for query recalibration loss computation. ~\textbf{Positive:} Only input Hungarian matched outputs for loss computation. We assign GT IoU scores to the unmatched queries by greedy matching to the GT boxes.}
\vspace{0.05in}
\resizebox{0.8\linewidth}{!}{%
\begin{tabular}{l|cccc}
\hline
Training strategies   & \multicolumn{1}{l}{$AP$} & \multicolumn{1}{l}{$AP_{50}$} & \multicolumn{1}{l}{$AP_{75}$} & $AR$ \\ \hline
Baseline &   22.3  &  41.5   &   21.2  &  51.4  \\
\hline
All & 22.5 & 38.7 & 22.6 & 49.4   \\
Positive &  25.9 &  44.5  &  26.1  &  53.9    \\ \hline
\end{tabular}}
\label{tab:iou_strategy}
\end{table}

\parsection{UDB10 Benchmark}
Table~\ref{tab:UDB10} shows the detailed results comparison between Faster R-CNN~\cite{ren2015faster}, DN-DETR~\cite{dndetr} and our Cascade-DN-DETR on the constructed UDB10 benchmark.
We compute UniAP as the mean of AP scores for each individual dataset component, where Cascade-DN-DETR obtains the highest 44.2 AP by improving the baseline performance for 5.7 AP and outperforms Faster R-CNN by 2.9 AP under the same R-50 backbone. The significant advancements reveal the generalizability of our approaches, without requiring any domain adaptation designs. 

For the six task-specific and small-scale datasets in UDB10, we further compare model finetuning results by taking their corresponding COCO pretrained model as initialization. We find that the result of Faster R-CNN with COCO pretraining only has a slight increase in most dataset components. However, the COCO finetuning is much more crucial for DETR-based approaches. For example, with COCO initialization, the AP$_{75}$ of DN-Def-DETR on Paintings~\cite{painting} improves drastically from 1.2 to 19.9, while Cascade-DN-Def-DETR boost from 9.0 to 21.5. However, Cascade-DN-Def-DETR still consistently outperforms the strong baseline DN-Def-DETR on all dataset components.

\parsection{Convergence Speed Comparison}
In Figure~\ref{fig:converge_speed}, we provide the convergence speed comparison on four task-specific benchmarks UVO~\cite{uvo}, Cityscapes~\cite{cityscapes}, Brain tumor~\cite{tumor} and Documentparts~\cite{document}. Note that DN-DETR has already been significantly speed up by its denoising branch during training. Our Cascade-DN-DETR outperforms the strong baseline DN-DETR across all datasets by a significant margin at various training stages, and converges much faster.

\begin{table}[t]
\centering%
\caption{Comparison with SOTA methods on COCO \textit{val2017}. All comparing methods are trained for 12 epochs. Asterisked models (*) were trained by ourselves. \textbf{Def}: deformable. We implement DAB-Deformable-DETR by removing the dn part in DN-Deformable-DETR.}
\vspace{0.05in}
\resizebox{1.0\linewidth}{!}{%
\begin{tabular}{@{}l@{~}c@{~~}c@{~~}c@{~~}c@{~~}c@{~~}c@{~~}c@{~~}c@{~~}c@{}}
\toprule
Model           & Base  & \multicolumn{1}{c}{$AP$} & \multicolumn{1}{c}{$AP_{50}$} & \multicolumn{1}{c}{$AP_{75}$} & \multicolumn{1}{l}{$AP_S$} & \multicolumn{1}{l}{$AP_M$} & \multicolumn{1}{l}{$AP_L$} & \multicolumn{1}{l}{Params} \\ \midrule
Faster-RCNN~\cite{ren2015faster}           &    R50      &  37.9   &  58.8  &   41.1    &   22.4   &  41.1    &  49.1 & 40M   \\
Cascade-RCNN~\cite{cai2018cascade}           &    R50      &  40.4  &  58.9  &   44.1    &   22.8   &  43.7    &  54.0  & 69M  \\
DETR~\cite{carion2020detr}            &    R50        &  15.5  &   29.4    &  14.5     &  4.3    &   15.1   &  26.7  & 41M  \\
Def DETR~\cite{zhu2020deformable} &    R50        &  37.2  &   55.5    &   40.5    &   21.1   &  40.7    &  50.5 & 40M   \\
CondDETR~\cite{meng2021conditional} &    R50       & 32.0 & 51.8  &  33.5  &  14.1 &  34.7 &  47.9  & 43M \\   
DAB-DETR(DC5)~\cite{dabdetr} &    R50       &  38.0   &  60.3     &   39.8    &  19.2    &  40.9    &  55.4  & 44M  \\
DE-CondDETR~\cite{dedetr}         &     R50      &  35.6  &  55.2  &  37.8  &  20.6  & 38.5 & 48.3 & 44M\\\midrule
DN-Def-DETR~\cite{dndetr}        &    R50       &  43.4  &  61.9     &  47.2     &  24.8    &  46.8    &  59.4 & 48M    \\
\textbf{Cascade-DN-Def-DETR}            &    R50   &   45.5$_{\uparrow\textbf{2.1}}$ &   62.2$_{\uparrow\textbf{0.3}}$    &   49.4$_{\uparrow\textbf{2.2}}$    &  27.3    &   50.0   &   62.4   & 48M \\ \midrule
DINO*~\cite{dino}          &    R50       &  48.8 & 66.2 & 53.1 & 31.1 & 52.0 & 63.0 & 47M  \\
\textbf{Cascade-DINO}         &    R50      &  49.7$_{\uparrow\textbf{0.9}}$  & 67.1$_{\uparrow\textbf{0.9}}$   &  54.1$_{\uparrow\textbf{1.0}}$  &  32.4  &  53.5 &  65.1   & 48M \\ \midrule
DAB-Def-DETR*~\cite{dabdetr}        &    R101  & 37.1 & 55.6 & 40.0 &19.3& 41.2 & 51.6 & 67M  \\
\textbf{Cascade-DAB-Def-DETR }        &    R101   & 42.7$_{\uparrow\textbf{5.6}}$ & 60.0$_{\uparrow\textbf{4.4}}$ & 46.3$_{\uparrow\textbf{6.3}}$ & 24.5 &47.7&58.4 &  67M  \\ \midrule
DN-Def-DETR~\cite{dndetr}        &    R101      &  44.1 &  62.8  &  47.9 & 26.0  & 47.8 &  61.3  & 67M  \\
\textbf{Cascade-DN-Def-DETR }          &    R101   &  46.5$_{\uparrow\textbf{2.4}}$  &  63.7$_{\uparrow\textbf{0.9}}$ &  50.4$_{\uparrow\textbf{2.5}}$ &   27.5 & 50.6 &  63.8  &67M \\ %\midrule
\bottomrule
\end{tabular}}
\label{tab:coco}
\vspace{-0.1in}
\end{table}

\begin{table}[t]
\centering
\caption{State-of-the-art results comparison on UVO~\cite{uvo}. All comparing methods are trained for 12 epochs. Both Cascade-DN-Def-DETR and Cascade-DAB-Def-DETR significantly surpass their strong baselines for over 7.0 AP$_{75}$.}
\vspace{0.05in}
\resizebox{1.0\linewidth}{!}{%
\begin{tabular}{lcccccccc}
\hline
Model & \multicolumn{1}{l}{Base} & \multicolumn{1}{c}{$AP$} & \multicolumn{1}{c}{$AP_{50}$} & \multicolumn{1}{c}{$AP_{75}$} & \multicolumn{1}{l}{$AP_S$} & \multicolumn{1}{l}{$AP_M$} & \multicolumn{1}{l}{$AP_L$} & \multicolumn{1}{l}{$AR$} \\ \hline
Faster-RCNN~\cite{ren2015faster}           &    R50     &  24.7  & 48.4  & 22.1 &  11.1 &  21.0  &  32.9  &  43.6  \\
Def DETR~\cite{zhu2020deformable} &    R50        & 17.7 & 34.9 & 16.3  & 6.7 & 15.8  & 25.3  &  51.4  \\ 
DE-CondDETR~\cite{dedetr}        &     R50     & 17.4 & 32.1 &  16.4 &  7.5 & 14.0  & 25.6  &  52.5 \\ \hline
DAB-Def-DETR~\cite{dabdetr}         &    R101    &  20.0  &  38.8     &  18.9     &   7.4   &   17.0   &  28.2    &  50.4  \\
\textbf{Cascade-DAB-Def-DETR}         &    R101      & 27.3$_{\uparrow\textbf{7.3}}$  & 44.1$_{\uparrow\textbf{5.3}}$   &  27.6$_{\uparrow\textbf{8.7}}$ & 11.0  & 23.1  & 37.7  &  58.0 \\ \hline
DN-Def-ETR~\cite{dndetr}         &    R50       &  22.3  &  41.5      &  21.2     &   7.1   &  16.9    &  33.1    &  51.4  \\
\textbf{Cascade-DN-Def-DETR}         &    R50       &  28.4$_{\uparrow\textbf{6.1}}$  &   44.9$_{\uparrow\textbf{3.4}}$    &   28.7$_{\uparrow\textbf{7.5}}$    &   10.7   &   22.5   &   40.7   &  58.2  \\ \hline
\end{tabular}}
\label{tab:uvo}
\end{table}

\begin{table}[]
\centering
\caption{State-of-the-art results comparison on Cityscapes~\cite{cityscapes}. Both Cascade-DN-Def-DETR and Cascade-DAB-Def-DETR achieve over 10.0 AP$_{75}$ performance gain over their counterparts.} 
\resizebox{1.0\linewidth}{!}{%
\begin{tabular}{lccccccccc}
\hline
Model           & \multicolumn{1}{l}{Base} & \multicolumn{1}{l}{Epoch} & \multicolumn{1}{c}{$AP$} & \multicolumn{1}{c}{$AP_{50}$} & \multicolumn{1}{c}{$AP_{75}$} & \multicolumn{1}{l}{$AP_S$} & \multicolumn{1}{l}{$AP_M$} & \multicolumn{1}{l}{$AP_L$}  \\ \hline
Faster-RCNN~\cite{ren2015faster}           &    R50      &   36   &  30.1  & 53.2 &  30.3 &  8.5  &  31.2 &  51.0   \\
DETR~\cite{carion2020detr}            &    R50      &   300   &  11.5  &   26.7    &  8.6     &   2.5   &   9.5   &  25.1      \\
Def DETR~\cite{zhu2020deformable} &    R50      &   50   &   27.3  &   49.2    &   26.3    &   8.7   &    28.2  &   45.7     \\ 
CondDETR~\cite{meng2021conditional}        &     R50     &   50   &   12.1  &   28.0    &   9.1   &   2.2   &   9.8   &  27.0        \\
DE-CondDETR~\cite{dedetr}        &     R50     &   50   &   26.8  &   47.8    &   25.4    &   6.8   &   25.6   &  46.6   \\\hline
DAB-Def-DETR~\cite{dabdetr}         &    R101    &  50  & 17.3  &  34.5 & 15.0 & 4.1 & 17.9 &  32.3\\
\textbf{Cascade-DAB-Def-DETR}        &    R101      &  50 & 25.4$_{\uparrow\textbf{8.1}}$  & 43.9$_{\uparrow\textbf{9.4}}$ & 25.0$_{\uparrow\textbf{10.0}}$ & 6.7 & 25.8 & 46.4\\ \hline
DN-Def-DETR~\cite{dndetr}         &    R50      &   50   &  19.0  &  39.3      &  15.7     &   4.9   &  19.8    &  35.5     \\
\textbf{Cascade-DN-Def-DETR}         &    R50      &   50   & 29.0$_{\uparrow\textbf{10.0}}$  &   49.5$_{\uparrow\textbf{10.2}}$    &    28.4$_{\uparrow\textbf{12.7}}$    &   9.1    &   28.4   &   51.5       \\ \hline
\end{tabular}}
\label{tab:cityscapes}
\end{table}

\begin{table}[]
\centering
\caption{Detailed results comparison on the proposed UDB10 benchmark using R50 backbone. All 
methods are initialized from ImageNet pretrained model. We take DN-Def-FETR as the strong baseline to build our Cascade-DN-Def-FETR.}
\vspace{0.05in}
\resizebox{1.0\linewidth}{!}{%
\begin{tabular}{l|ccc|ccc|ccc}
\hline
 &  \multicolumn{3}{c|}{Faster RCNN~\cite{ren2015faster}}    & \multicolumn{3}{c|}{DN-Def-DETR~\cite{dndetr}} & \multicolumn{3}{c}{\textbf{Cascade-DN-Def-DETR}} \\ \hline
Dataset  & $AP$ & $AP_{50}$ & $AP_{75}$  & $AP$ & $AP_{50}$ & $AP_{75}$  & $AP$ & $AP_{50}$ &  $AP_{75}$\\ \hline
COCO~\cite{lin2014microsoft}  &  37.9  & 58.8 & 41.1 &   43.4   &  61.9    &  47.2  & 45.5$_{\uparrow\textbf{2.1}}$ & 62.2$_{\uparrow\textbf{0.3}}$ & 49.4$_{\uparrow\textbf{2.2}}$  \\ 
UVO~\cite{uvo}  &  24.7  & 48.4 & 22.1 &   22.3   &  41.5    &  21.2  & 28.4$_{\uparrow\textbf{6.1}}$ & 44.9$_{\uparrow\textbf{3.4}}$ & 28.7$_{\uparrow\textbf{7.5}}$ \\ 
Cityscapes~\cite{cityscapes}  &  30.1  & 53.2 & 30.3 &  19.0   &  39.3    &  15.7  & 29.0$_{\uparrow\textbf{10.0}}$ & 49.5$_{\uparrow\textbf{10.2}}$ & 28.4$_{\uparrow\textbf{12.7}}$    \\ 
BDD100K~\cite{bdd100k} & 31.0 & 55.9 & 29.4 & 28.2 & 53.9 & 24.8 & 30.2$_{\uparrow\textbf{2.0}}$  & 55.0$_{\uparrow\textbf{1.1}}$  & 27.9$_{\uparrow\textbf{3.1}}$  \\
Brain tumor~\cite{tumor} &  43.5  & 75.1 & 45.0 &   38.9   &  71.6    &  38.6  & 46.5$_{\uparrow\textbf{7.6}}$ & 75.6$_{\uparrow\textbf{4.0}}$ & 49.4$_{\uparrow\textbf{10.8}}$ \\ 
Document~\cite{document}  &   48.0  & 66.2 &  55.6 &  44.7  &  64.1  & 50.3 & 50.9$_{\uparrow\textbf{16.2}}$ & 66.6$_{\uparrow\textbf{2.5}}$ & 58.4$_{\uparrow\textbf{8.1}}$   \\ 
Smoke~\cite{smoke} &  67.1  &  92.9 &  80.6 & 66.5   &   91.6   &  77.8 & 71.8$_{\uparrow\textbf{5.3}}$ & 91.9$_{\uparrow\textbf{0.3}}$ & 82.9$_{\uparrow\textbf{5.1}}$ \\
EgoHands~\cite{smoke} &  74.9  & 96.9 & 90.4 &  74.4    & 97.4  & 89.2   & 77.6$_{\uparrow\textbf{3.3}}$  &  98.3$_{\uparrow\textbf{0.9}}$ & 91.5$_{\uparrow\textbf{2.3}}$   \\
PlantDoc~\cite{plantdoc}   &  38.9  &  60.8  &  44.9 & 45.0 & 61.7 & 53.7 & 49.1$_{\uparrow\textbf{4.1}}$ & 63.9$_{\uparrow\textbf{2.2}}$ & 56.5$_{\uparrow\textbf{2.8}}$\\
Paintings~\cite{painting}   & 17.0 & 50.1 & 6.3 & 2.2 & 5.8 & 1.2 & 13.4$_{\uparrow\textbf{11.2}}$ & 33.1$_{\uparrow\textbf{27.3}}$ & 9.0$_{\uparrow\textbf{7.8}}$   \\  \hline
\textbf{UniAP}   &  41.3 &  &  & 38.5 &  &  & \textbf{44.2}$_{\uparrow\textbf{5.7}}$ & &    \\  \hline
\end{tabular}}
\label{tab:UDB10}
\end{table}

\begin{table}[]
\centering
\caption{Finetuning results for the six small-scale task-specific datasets in our UDB10 benchmark using R50 backbone. All  finetuned methods are from their COCO pretrained model. }
\vspace{0.05in}
\resizebox{1.0\linewidth}{!}{%
\begin{tabular}{l|ccc|ccc|ccc}
\hline
 &  \multicolumn{3}{c|}{Faster RCNN~\cite{ren2015faster}}    & \multicolumn{3}{c|}{DN-Def-DETR~\cite{dndetr}} & \multicolumn{3}{c}{\textbf{Cascade-DN-Def-DETR}} \\ \hline
Dataset  & $AP$ & $AP_{50}$ & $AP_{75}$  & $AP$ & $AP_{50}$ & $AP_{75}$  & $AP$ & $AP_{50}$ &  $AP_{75}$\\ \hline
Brain tumor~\cite{tumor} & 44.2 & 74.1 &46.3 &48.0 & 78.1 & 51.7 & 51.5$_{\uparrow\textbf{3.5}}$ & 79.6 & 56.0\\ 
Document~\cite{document}  &49.7 & 67.7& 58.0& 51.6 & 67.7 & 60.8 & 52.7$_{\uparrow\textbf{1.1}}$ & 68.0 & 61.1   \\ 
Smoke~\cite{smoke} &68.3 & 90.3 & 82.4& 72.9 & 94.9 & 86.0 & 74.3$_{\uparrow\textbf{1.4}}$ & 93.7 &86.8\\
EgoHands~\cite{egohands} & 75.8 & 96.9 & 90.0 & 77.4 & 98.7 & 92.1 & 79.4$_{\uparrow\textbf{2.0}}$ & 97.9 & 93.4    \\
PlantDoc~\cite{plantdoc} & 37.5 & 57.3 & 39.9 & 50.2 &62.9 & 57.3 & 52.7$_{\uparrow\textbf{2.5}}$ & 66.5 & 61.1  
\\
Paintings~\cite{painting}   & 24.1 & 59.5 & 15.4& 28.6 & 66.3 &19.9 & 29.4$_{\uparrow\textbf{0.8}}$ & 66.0 & 21.5  \\ \hline
\end{tabular}}
\label{tab:UDB10_pretrain}
\vspace{-0.2in}
\end{table}

\section{Conclusion}
We present \modelname, the first DETR-based detector targeting for high-quality universal detection. To benefit future research on 
universal detection, we  propose a large-scale universal object detection benchmark UDB10, which is composed of 10 subdatasets from various real-life domains. Injected with local object-centric prior, \modelname achieves significant advantages in a wide range of detection applications, especially in higher IoU thresholds. We hope the detection community to focus more on real-life and practical applications when evaluating the detector performance, not only considering the \textit{de facto} COCO, especially for the data-sensitive DETR-based approaches.

\section{Appendix}
In this supplementary material, we first present the comprehensive quantitative results of our Cascade-DN-DETR and Cascade-DINO, and compare them to state-of-the-art methods on UDB10 benchmark in Section~\ref{sec:supp_exp}.
We also provide detailed training convergence curve comparison, more study on deformable attention and alternative training manners.
In Section~\ref{sec:supp_vis}, we then show qualitative results comparisons of our Cascade-DN-DETR to DN-DETR~\cite{dndetr}, including localization quality, score re-calibration and occlusion cases.
Finally, we provide more implementation/training details in~Section~\ref{sec:supp_details}.
\subsection{Supplementary Experiments}
\label{sec:supp_exp} 
\vspace{0.1in}
\parsection{More Results Comparison on UDB10} In Table~\ref{tab:UDB10_supp}, we provide comprehensive and detailed experiments results comparison on all 10 dataset components of the constructed UDB10. The compared six methods include Faster R-CNN~\cite{ren2015faster}, Cascade R-CNN~\cite{cai2018cascade}, DN-Def-DETR~\cite{dndetr}, Cascade-DN-Def-DETR (Ours), the most recent DINO~\cite{dino} and Cascade-DINO (Ours). Cascade-DINO achieves the best UniAP 45.0 and UniAP$_{75}$ 49.0 among all comparing methods. It's worth to mention that both Cascade-DINO and Cascade-DN-Def-DETR boost the performance of their strong baselines on all 10 dataset components consistently. This shows the generalizability and effectiveness of our proposed cascade attention and IoU-aware query re-calibration. Interestingly, we observe that although DINO obtains over 3.3 AP advantage over Cascade-DN-Def-DETR on COCO dataset, its UniAP 0.6 point lower than the our Cascade-DN-Def-DETR (43.6 \vs 44.2). This indicates that the robustness of the most recent DINO~\cite{dino} across domains still have improvement space.

\begin{table*}[]
\centering
\caption{Quantative Results Comparison on the constructed UDB10 benchmark using R50 backbone. All 
methods are initialized from ImageNet pretrained model. We take DN-Def-DETR~\cite{dndetr} as the baseline to build our Cascade-DN-Def-DETR. We also take DINO~\cite{dino} as a stronger baseline, replacing the deformable transformer decoder with our cascade transformer decoder and building our Cascade-DINO. The UniAP metric computes the mean of AP for each individual dataset component.}
\vspace{0.05in}
\resizebox{1.0\linewidth}{!}{%
\begin{tabular}{l|ccc|ccc|ccc|ccc|ccc|ccc}
\hline
 &  \multicolumn{3}{c|}{Faster RCNN~\cite{ren2015faster}}    & \multicolumn{3}{c|}{DN-Def-DETR~\cite{dndetr}} & \multicolumn{3}{c|}{\textbf{Cascade-DN-Def-DETR}}  &  \multicolumn{3}{c|}{Cascade RCNN~\cite{cai2018cascade}}    & \multicolumn{3}{c|}{DINO~\cite{dino}} & \multicolumn{3}{c}{\textbf{Cascade-DINO}} \\ \hline
Dataset  & $AP$ & $AP_{50}$ & $AP_{75}$  & $AP$ & $AP_{50}$ & $AP_{75}$  & $AP$ & $AP_{50}$ &  $AP_{75}$ & $AP$ & $AP_{50}$ & $AP_{75}$  & $AP$ & $AP_{50}$ & $AP_{75}$  & $AP$ & $AP_{50}$ & $AP_{75}$ \\ \hline
COCO~\cite{lin2014microsoft}  &  37.9  & 58.8 & 41.1 &   43.4   &  61.9    &  47.2  & 45.5$_{\uparrow\textbf{2.1}}$ & 62.2 & 49.4 & 40.4 & 58.9 &  44.1 &  48.8 &  66.2 & 53.1 & 49.7$_{\uparrow\textbf{0.9}}$  & 67.1 &54.1 \\ 
UVO~\cite{uvo}  &  24.7  & 48.4 & 22.1 &   22.3   &  41.5    &  21.2  & 28.4$_{\uparrow\textbf{6.1}}$ & 44.9 & 28.7 &  26.2 & 46.7 & 25.2 &  30.2  &  46.9 & 30.5 & 32.7$_{\uparrow\textbf{2.5}}$ & 50.2 & 33.4 \\ 
Cityscapes~\cite{cityscapes}  &  30.1  & 53.2 & 30.3 &  19.0   &  39.3    &  15.7  & 29.0$_{\uparrow\textbf{10.0}}$ & 49.5 & 28.4 &  31.8 & 54.4 & 30.9 & 34.5  & 56.6 & 34.5 & 34.8$_{\uparrow\textbf{0.3}}$ & 57.3 & 33.7    \\ 
BDD100K~\cite{bdd100k} & 31.0 & 55.9 & 29.4 & 28.2 & 53.9 & 24.8 & 30.2$_{\uparrow\textbf{2.0}}$  & 55.0  & 27.9  & 32.4 & 56.3 & 31.6 & 34.4 & 60.7 & 32.7 & 35.6$_{\uparrow\textbf{1.2}}$ & 61.8 &34.0 \\
Brain tumor~\cite{tumor} &  43.5  & 75.1 & 45.0 &   38.9   &  71.6    &  38.6  & 46.5$_{\uparrow\textbf{7.6}}$ & 75.6 & 49.4 & 46.2 & 74.2 & 49.6 & 46.4 & 76.8 & 49.1 & 48.6$_{\uparrow\textbf{2.2}}$ & 77.8 & 52.2\\ 
Document~\cite{document}  &   48.0  & 66.2 &  55.6 &  44.7  &  64.1  & 50.3 & 50.9$_{\uparrow\textbf{16.2}}$ & 66.6 & 58.4  & 50.3& 66.3 & 58.9 & 47.7 & 63.2 & 55.9 & 49.6$_{\uparrow\textbf{1.9}}$ & 65.8 &58.1\\ 
Smoke~\cite{smoke} &  67.1  &  92.9 &  80.6 & 66.5   &   91.6   &  77.8 & 71.8$_{\uparrow\textbf{5.3}}$ & 91.9 & 82.9 & 70.4  & 91.3 & 83.5 &  69.4 & 92.4 & 80.7 & 69.7$_{\uparrow\textbf{0.3}}$ & 92.6 &80.4\\
EgoHands~\cite{smoke} &  74.9  & 96.9 & 90.4 &  74.4    & 97.4  & 89.2   & 77.6$_{\uparrow\textbf{3.3}}$  &  98.3 & 91.5  & 76.4 & 96.9 & 91.5 & 77.7 & 97.9 & 91.8 & 78.0$_{\uparrow\textbf{0.3}}$ & 98.0 & 91.6   \\
PlantDoc~\cite{plantdoc}   &  38.9  &  60.8  &  44.9 & 45.0 & 61.7 & 53.7 & 49.1$_{\uparrow\textbf{4.1}}$ & 63.9 & 56.5 & 37.5 & 55.3 & 43.6 & 35.1 & 49.7 & 39.9 & 38.3$_{\uparrow\textbf{3.2}}$ & 53.8 & 44.2\\
Paintings~\cite{painting}   & 17.0 & 50.1 & 6.3 & 2.2 & 5.8 & 1.2 & 13.4$_{\uparrow\textbf{11.2}}$ & 33.1 & 9.0  & 18.0 & 50.7 & 8.1 & 12.0  & 30.3 & 6.7 & 13.4$_{\uparrow\textbf{1.4}}$ & 34.3 &7.9 \\  \hline
\textbf{UniAP}   &  41.3 &  &  & 38.5 &  &  & \textbf{44.2}$_{\uparrow\textbf{5.7}}$ & &     & 43.0  &  &  & 43.6 &  &  & \textbf{45.0}$_{\uparrow\textbf{1.4}}$ & &   \\  \hline
\textbf{UniAP$_{75}$}  & &   & 44.6 &  &  & 42.0 &  &  & \textbf{48.2}$_{\uparrow\textbf{6.2}}$  & &   & 46.7 &  &  & 47.5 &  &  & \textbf{49.0}$_{\uparrow\textbf{1.5}}$   \\  \hline
\end{tabular}}
\label{tab:UDB10_supp}
\end{table*}

\vspace{0.1in}

\begin{figure*}[!t]
\centering
\vspace{-0.3in}
\includegraphics[width=1.0\linewidth]{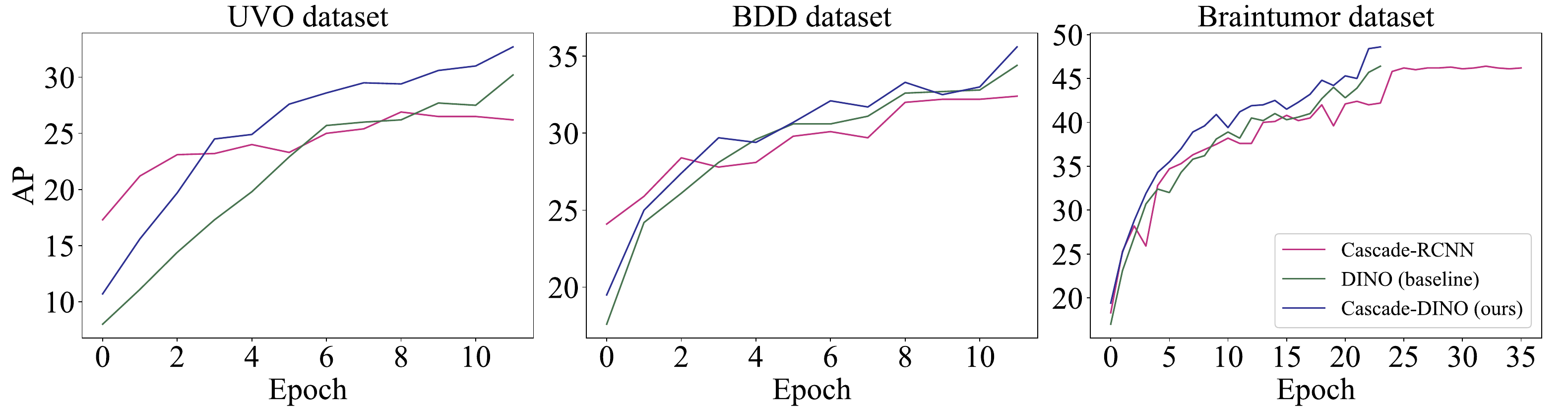}
\caption{Quantitative results comparison between Cascade R-CNN~\cite{cai2018cascade}, DINO~\cite{dino} (Baseline) and Cascade-DINO (Ours) per training epoch on UVO~\cite{uvo}, BDD~\cite{bdd100k}, Brain tumor~\cite{tumor}. These datasets cover three various detection application domains. Cascade-DINO achieves stable performance growth during training, and outperforms Cascade R-CNN and DINO consistently on all three domains. Note that DINO is a very recent work which has already been significantly speed up by the usage of denoising branch and two-stage training.}
\vspace{-0.1in}
\label{fig:converge_speed_supp}
\end{figure*}

\parsection{Convergence Speed Comparison} 
In Figure~\ref{fig:converge_speed_supp}, we show the detection results comparison between Cascade R-CNN~\cite{cai2018cascade}, DINO~\cite{dino} (Baseline) and Cascade-DINO (Ours) per training epoch.
We sample three dataset components UVO~\cite{uvo}, BDD~\cite{bdd100k} and Braintumor~\cite{tumor} out of UDB10, which belong to various domains in open-world, self-driving and medical analysis. 
On UVO and BDD datasets, we find that Cascade R-CNN converges faster in the first three epochs, but its performance is quickly saturated and surpassed by our Cascade-DINO after training for 3 epochs. 
Cascade-DINO achieves stable performance growth during training, and outperforms Cascade R-CNN and DINO consistently in all three domains.
\vspace{0.1in}

\parsection{Comparison to Deformable Attention}  We compare Cascade attention (CA) with Deformable attention (DA) in Tab.~\ref{tab:vs_deformable}, where merely replacing the Deformable attention in DINO to Cascade attention (\ie \textbf{w/o} QR) promotes the results from 30.2 to 31.9 on UVO.

\parsection{Comparison to Box-constrained Deformable Attention} In Table~\ref{tab:constrained_deform}, we compare our cascade attention to both the standard deformable attention and box-constrained deformable attention. 
We design box-constrained deformable attention by normalizing its learnable offsets within the predicted bounding boxes.
Box-constrained deformable attention slightly increases the result of standard deformable attention by 0.8 AP. However, its performance is still 1.7 AP lower than our cascade-attention. This gap may due to the insufficient/limited sampling points of deformable attention in the box regions.
\vspace{0.05in}

\parsection{Adopting GT boxes in the Initial Training Stage} We try an alternative training manner to replace the predicted boxes with GT boxes for constraining attention in the initial training stage. 
The intuition is that at the beginning stage, the predicted boxes by learnable queries are inaccurate, which can be replaced by the corresponding GT box via greedy matching.
We take Cascade-DINO as baseline and set the ratio of using GT boxes at the first iteration being 25\%. 
This ratio then linearly decreases to 0 at the last training iteration after 12 epochs.
However, we find it leads to 3.0 AP performance decrease (32.7 $\rightarrow$ 29.7) on UVO dataset comparing to using predicted boxes for the whole network training stage.

\subsection{More Qualitative Comparisons}
\label{sec:supp_vis}

\begin{figure*}[!t]
\centering
\includegraphics[width=1.0\linewidth]{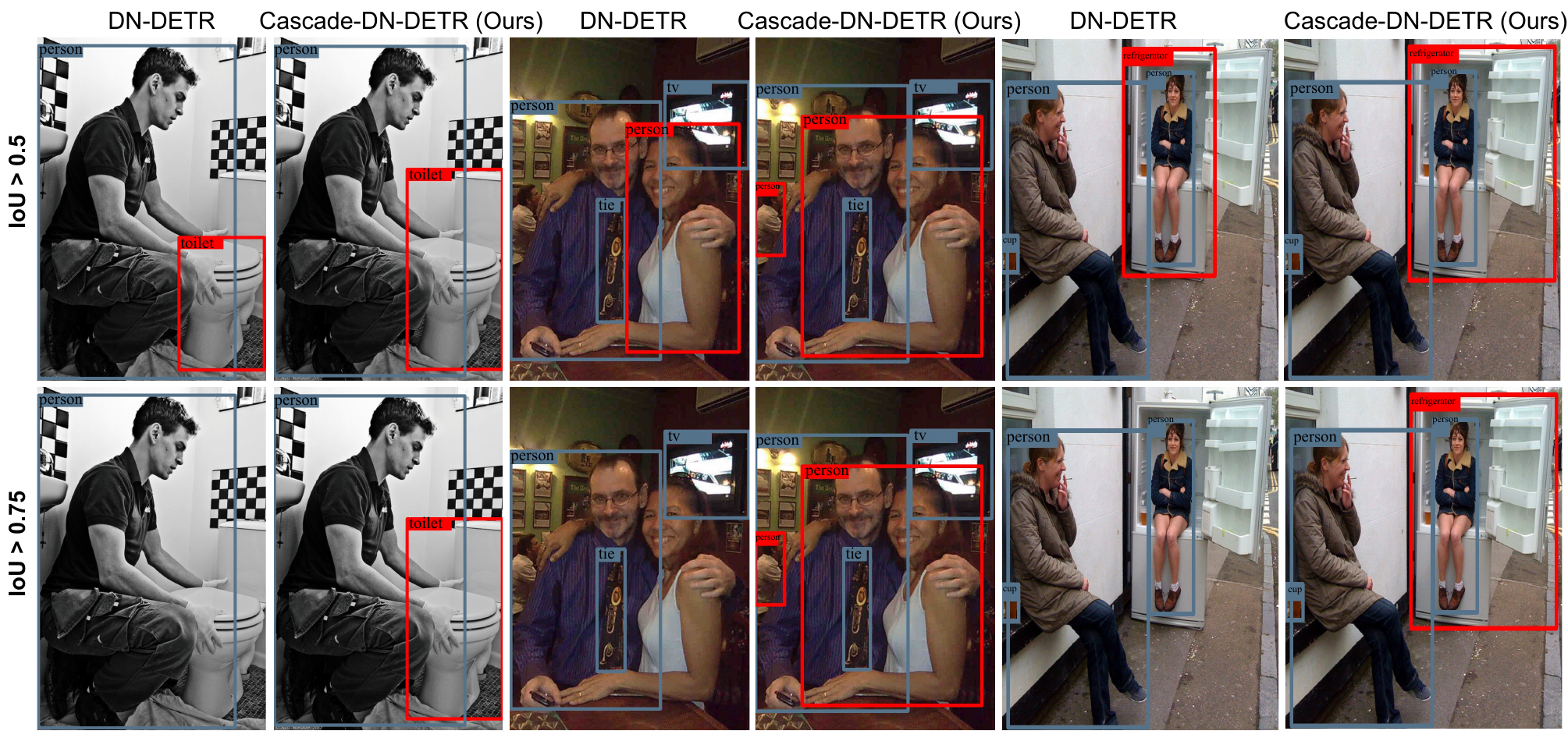}
\vspace{-0.2in}
\caption{Qualitative results comparison between DN-DETR~\cite{dndetr} and our Cascade-DN-DETR on COCO validation set. The box prediction differences are highlighted in red color. The first row shows box predictions satisfying IoU thresold (to GT box) larger than 0.5 while the bottom row shows IoU thresold larger than strict IoU (to GT) threshold 0.75. Taking the first column as example, the detected `toilet' object by baseline DN-DETR is filtered when using strict threshold 0.75.
}
\vspace{-0.05in}
\label{fig:high_quality}
\end{figure*}

\parsection{Visualization Comparison in Box Quality }
In Figure~\ref{fig:high_quality}, we compare the predicted box quality between baseline DN-DETR and our Cascade-DN-DETR. On COCO validation set, we visualize the predicted boxes satisfying IoU (to GT) thresholds 0.5 and 0.75 in the first row and second row respectively. The predicted boxes differences are highlighted in red. The detected lower-quality predictions by the baseline, such as `toilet' (left example) and `refrigerator' (right example) are highlighted in the red boxes.

\begin{figure*}[!t]
\centering
\vspace{-0.1in}
\includegraphics[width=0.85\linewidth]{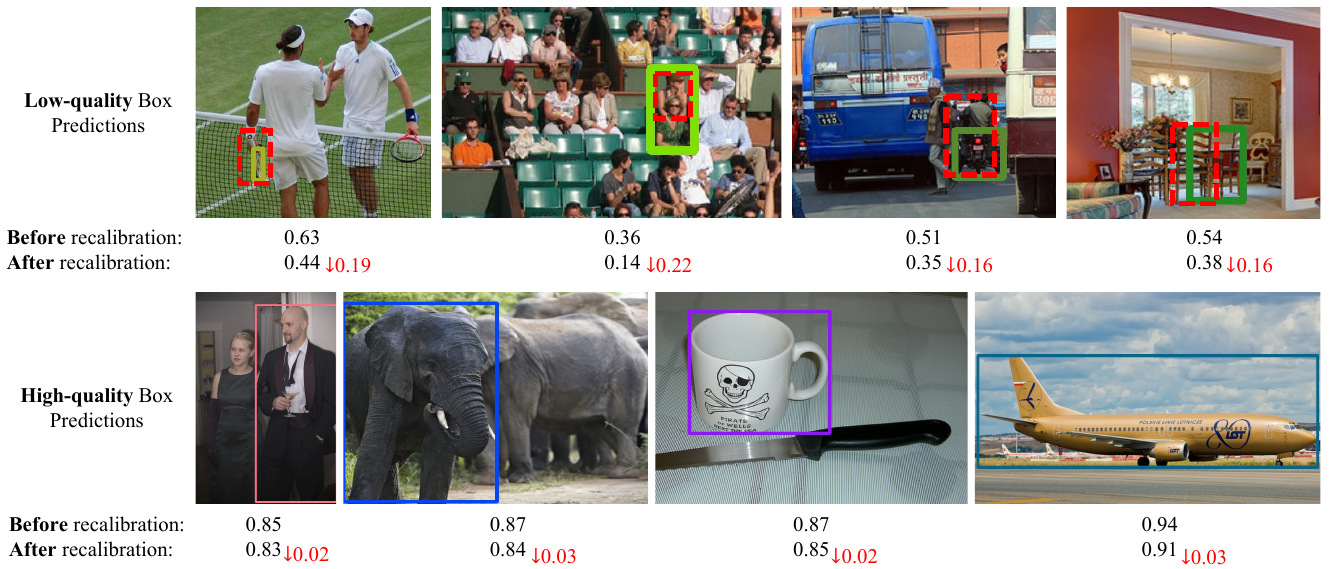}
\caption{Predicted boxes and corresponding scores of Cascade-DN-DETR before and after IoU-aware query re-calibration. In the first row, we visualize both the box prediction by our Cascade-DN-DETR and the corresponding GT boxes (in dotted line of red color). The first row shows that for low-quality predicted boxes (with small IoUs to the GT boxes), their confidence scores after re-calibration will have an obvious decrease to align with the low localization quality. The second row shows that for high-quality box predictions with high IoUs to GT boxes (not shown here due to overlapping), the re-calibration has a negligible influence on the original classification score. 
}
\label{fig:rescore}
\end{figure*}

\parsection{Visualization of Score Recalibration}
In Figure~\ref{fig:rescore}, we visualize the predicted boxes and corresponding confidence score before and after IoU-aware query re-calibration. For the low-quality box predictions with small IoUs to GT, their confidence scores typically have an obvious decrease of around 0.2. However, for the high-quality boxes (2nd row), the re-calibration has minor influences (around 0.02) on the predicted box scores. The recalibration adjusts the box confidence score to better reveal its localization quality. 

\begin{figure*}[!t]
\centering
\includegraphics[width=0.8\linewidth]{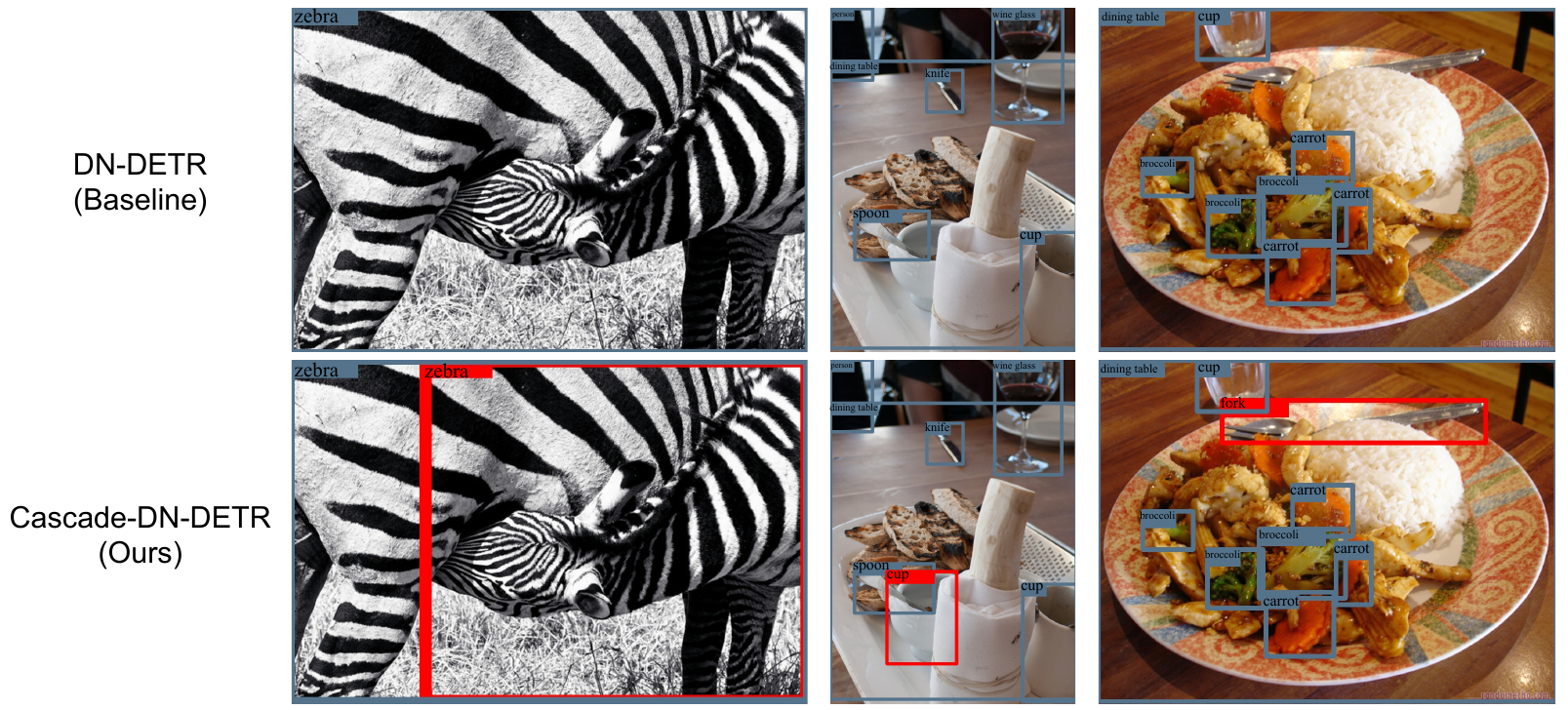}
\caption{Qualitative results comparison between DN-DETR~\cite{dndetr} and our Cascade-DN-DETR. The box prediction differences are highlighted in red. Comparing to DN-DETR (Baseline), our Cascade-DN-DETR can better detect objects for the challenging occlusion cases.
}
\label{fig:Occlusion}
\end{figure*}

\parsection{Visualization Comparison in Occlusion Cases} In Figure~\ref{fig:Occlusion}, comparing to DN-DETR, we find that Cascade-DN-DETR has a higher recall rate for detection in the challenging heavy occlusion cases. The box-constrained cascade attention can better attend to the target occluded objects with less distraction from their neighboring overlapping objects.

\begin{figure*}[!t]
\centering
\vspace{-0.1in}
\includegraphics[width=0.75\linewidth]{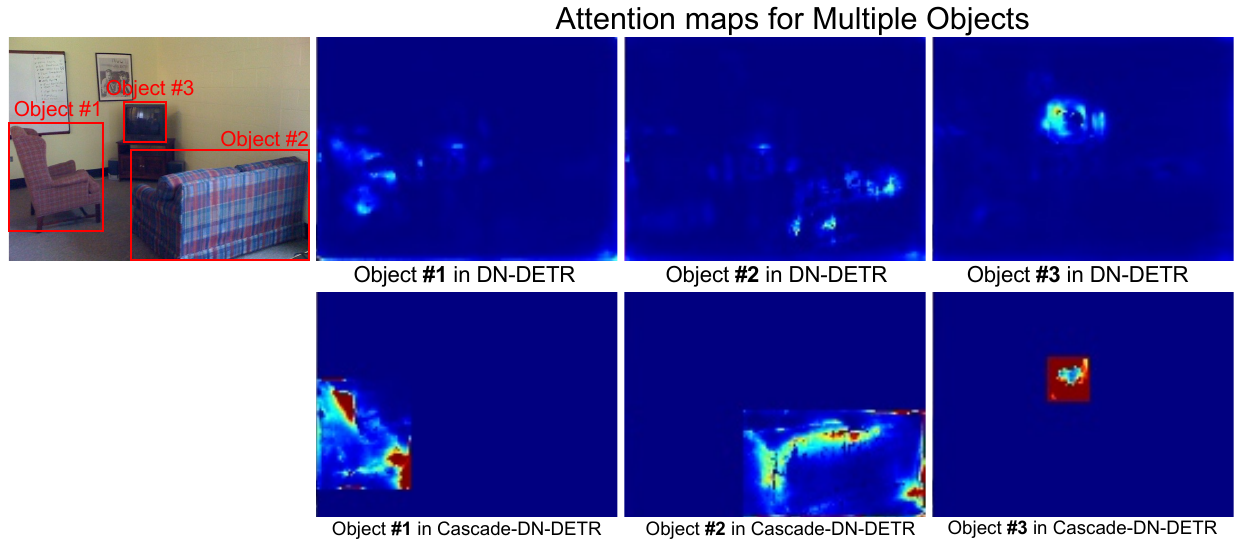}
\caption{
Visual comparison of cross-attention maps between DN-DETR~\cite{dndetr} and our Cascade-DN-DETR on COCO for multiple objects.}\vspace{-3mm}
\label{fig:multi_attn_map}%
\end{figure*}

\parsection{Visualization of Multi-object Attention Map} We provide cross-attention maps for multiple objects in Fig.~\ref{fig:multi_attn_map}. Cascade-DN-DETR's query attention focuses on the most relevant parts of the detected objects, while DN-DETR has a more scattered attention distribution.

\begin{table}[t]
\centering%
\caption{Cascade attention (CA) vs.\ Deformable attention (DA) on UVO. QR denotes Query Re-calibration.}
\resizebox{0.82\linewidth}{!}{%
\begin{tabular}{l|ccc|cccc}
\hline
Model  & DA & CA & QR & \multicolumn{1}{l}{$AP$} & \multicolumn{1}{l}{$AP_{50}$} & \multicolumn{1}{l}{$AP_{75}$} & $AR$  \\ \hline
DINO [46] & $\checkmark$  &  &  & 30.2 & 46.9 & 30.5 & 63.4 \\ \hline
 &$\checkmark$  &  &  $\checkmark$  & 31.5 & 47.9 & 32.1 & 63.4   \\
Cascade-DINO &  & $\checkmark$   & & 31.9 & 49.5 & 32.8 & 63.4   \\
  & & $\checkmark$ & $\checkmark$ & \textbf{32.7} & 50.2 & 33.4 & 63.0   \\ \hline
\end{tabular}}%
\vspace{-2mm}%
\label{tab:vs_deformable}
\end{table}

\begin{table}[t]
\centering
\caption{Ablation study on the box-constrained Deformable attention on UVO. \textbf{Baseline}: DINO with standard deformable attention in the transformer decoder. \textbf{Box-constrained deformable attention}: the learnable offsets around reference points are constrained inside the predicted boxes by normalization. Our Cascade-DINO adopts the cascade-attention.}
\vspace{0.1in}
\resizebox{1.0\linewidth}{!}{%
\begin{tabular}{l|cccc}
\hline
Model   & \multicolumn{1}{c}{$AP$} & \multicolumn{1}{l}{$AP_{50}$} & \multicolumn{1}{l}{$AP_{75}$} & $AR$ \\ \hline
DINO (Baseline, standard deformable attention) &   30.2  &  46.9   &   30.5  & 63.4   \\ \hline
Box-constrained Deformable attention & 31.0$_{\uparrow\textbf{0.8}}$ &47.2  &  31.7  &  62.8 \\ 
Casacde-DINO (Ours) &  32.7$_{\uparrow\textbf{2.5}}$ &  50.2  &  33.4  &  63.0    \\ \hline
\end{tabular}}
\label{tab:constrained_deform}
\end{table}

\subsection{More Implementation Details}
\label{sec:supp_details} 

\parsection{Implementation Details}
In the paper, we mainly adopt DN-DETR~\cite{dndetr} as our baseline. We use 4-scale feature maps with the help of a deformable encoder. For the transformer decoder, we apply cascade attention to each feature map and perform fusion in each layer. For query recalibration, we add one IoU head which is similar to the classification head but the output of the IoU head has only one channel for each query.  We employ 300 queries with one pattern to keep the same as 300 queries of traditional DETR methods. We perform the same Hungarian matching as traditional DETR-based methods and record the IoU of matched boxes and ground truth. We use this as the regression target for our IoU head. L2Loss is used for IoU regression. We adopt auxiliary losses and the same loss coefficients as other DETR methods. The loss coefficients for classification loss, box L1 loss, box giou loss and our recalibration loss are \{1.0, 5.0, 2.0, 2.0\}. For DAB-DETR~\cite{dabdetr}, we also use 4-scale feature maps and implement it by removing the dn-part of the previous DN-DETR.  For Faster-RCNN~\cite{ren2015faster} and Cascade R-CNN~\cite{cai2018cascade}, we use the standard implementation of mmdetection.

For all dataset components in the UDB10 benchmark, we use the default data augmentation of Faster-RCNN, Cascade-RCNN and DETR-based methods on COCO. Large datasets (more than 10k images) in the UDB10 benchmark are COCO, UVO, BDD100K, and EgoHands. For them, we train DN-DETR for 12 epochs with an initial learning rate  of $1\times10^{-5}$ for the backbone and $1\times10^{-4}$ for the transformer and drop the learning rate at the 10th epoch.   We use the AdamW optimizer with weight decay   $1\times10^{-4}$.  We train on 8 Nvidia GeForce RTX 3090 GPUs with a total batch size of 8. We train Faster-RCNN for 12 epochs and drop the learning rate at the 8th and 11th epochs. For other small datasets, we train DN-DETR for 50 epochs and drop the learning rate at the 40th epoch. We train Faster-RCNN for 36 epochs and drop the learning rate at the 24th and 33rd epochs. 

We also perform experiments on DINO~\cite{dino} using Resnet50 backbone. We use 900 queries for the DINO baseline and Cascade-DINO. We employ 4-scale feature maps from the deformable encoder. We implement Cascade-DINO by replacing its deformable decoder with our cascade decoder, which is the same as Cascade-DN-DETR.  Since DINO is a two-stage DETR detector, we also use the recalibrated score to select anchors from the transformer encoder.  DINO is a strong SOTA method and we adopt the same 12-epoch and 24-epoch training schedules as their paper. For large datasets (more than 10k images) in the UDB10 benchmark, we train Cascade-RCNN, DINO, and Cascade-DINO for 12 epochs. For other small datasets, we train Cascade-RCNN for 36 epochs and DINO, Cascade-DINO for 24 epochs. For Box-constrained deformable attention, we also use IoU recalibration for comparison with Cascade-DINO.

\parsection{Inference Details} During inference, we predict expected IoU scores (Eq.4 of the paper) for all queries and take them as the recalibrated scores as discussed in the paper. We perform cascade attention with the initial boxes and predicted boxes of each layer. We also add cascade attention and IoU recalibration to dn-part in training and do not use dn-part in inference, which is the same as the original DN-DETR. For the other inference settings, we kept the same with our baseline methods DN-DETR and DINO.

\clearpage
\clearpage

{\small
\bibliographystyle{ieee_fullname}
\bibliography{egbib}
}

\end{document}